\documentclass[11pt]{article}
\usepackage{acl}
\usepackage{times}
\usepackage{latexsym}
\usepackage[T1]{fontenc}
\usepackage[utf8]{inputenc}
\usepackage{microtype}
\usepackage{graphicx}
\usepackage{amsmath}
\usepackage{amssymb}
\usepackage{booktabs}
\usepackage{longtable}
\usepackage{multirow}
\usepackage{xcolor}
\usepackage{enumitem}
\usepackage{titlesec}
\titlespacing*{\section}{0pt}{8pt}{4pt}
\titlespacing*{\subsection}{0pt}{6pt}{3pt}
\titlespacing*{\paragraph}{0pt}{5pt}{4pt}
\usepackage{listings}
\usepackage{pdflscape}
\usepackage{tikz}
\usetikzlibrary{shapes.geometric, arrows.meta, positioning, fit, backgrounds, calc}
\usepackage{pgfplots}
\pgfplotsset{compat=1.17}
\usepackage{subcaption}
\usepackage{placeins}

\title{Reliability of LLM Judges for Evaluating Entity Alignment}

\author{
    Vaibhava Lakshmi Ravideshik$^1$ \and Mayank Kejriwal$^2$ \\
    $^1$ University of Michigan, Ann Arbor \\
    $^2$ Information Sciences Institute, University of Southern California \\
    \texttt{vlds@umich.edu, kejriwal@isi.edu}
  }

\begin{document}
\maketitle

\begin{abstract}
Entity Alignment (EA) identifies equivalent entities across knowledge
graphs and is critical for knowledge base integration and ontology
merging. Evaluating EA systems at scale requires expensive expert
annotation, making systematic assessment across diverse domains
practically infeasible. LLM-as-judge evaluation offers a potentially
scalable alternative, yet its reliability for structured prediction
tasks like EA remains unstudied. We present the first systematic
benchmarking study across three frontier models, three datasets, and
four EA systems, using perturbation bias diagnostics, meta-evaluation
across all dataset--judge--prompt combinations, and counterfactual
label-flip tests. We identify \emph{anchor bias}, a failure mode in
which judges invert discrimination when the system's decision label is
visible. Judge discrimination collapses under label-exposing prompts
(J-ROC-AUC 0.12--0.87), while a label-free protocol recovers
near-ceiling capability on distinctive-name datasets (0.98--1.00 for NeuSymEA)
and significant recovery on biomedical pairs (0.93--0.95).
Counterfactual label flips show that the visible label causally shifts
scores by more than 2 points in 53--99\% of pairs (FSR), with sensitivity
growing with model capability, a pattern we call the frontier model paradox:
stronger judges are more label-sensitive, not less, yet also better label-free
predictors. The direction of the shifts is mixed, so we report this as an
observed pattern. A blinded two-annotator human evaluation
(102 pairs, Cohen's $\kappa=0.902$) provides a gold-label reference.
We release the first biomedical EA
benchmark (MeSH--SNOMED~CT, 15K pairs) and a reproducible auditing
framework for LLM judge reliability in EA.
\textit{Code and data are available at \url{https://github.com/vaibhavalakshmiravideshik/llm-as-a-judge-entity-alignment}.}
\end{abstract}

\section{Introduction}\label{sec:intro}

Entity alignment determines when entities in different knowledge
graphs refer to the same real-world concept and is fundamental for
knowledge base integration, ontology merging, and scientific data
linking. Existing approaches range from translation-based embedding
methods \cite{bordes2013translating, sun2017cross, wang2018cross}
to hybrid neural-symbolic systems \cite{cheng-etal-2025-easyea,
chen2025neusymea}. Despite advances in EA methodology, a critical
bottleneck persists: evaluating whether predictions are correct
requires expensive expert annotation, making large-scale
benchmarking practically infeasible. LLM-as-judge evaluation can be
a promising scalable alternative \cite{zheng2023judging,
fu2023gptscore}, yet a troubling pattern is emerging: LLMs exhibit
systematic vulnerabilities when asked to make precise judgments.
Recent work shows that a single irrelevant clause causes performance
drops of up to 65\% \cite{mirzadeh2025gsmsymbolic}, suggesting LLMs
replicate training patterns rather than reason genuinely.

\begin{figure}[t]
\centering
\includegraphics[width=0.70\columnwidth]{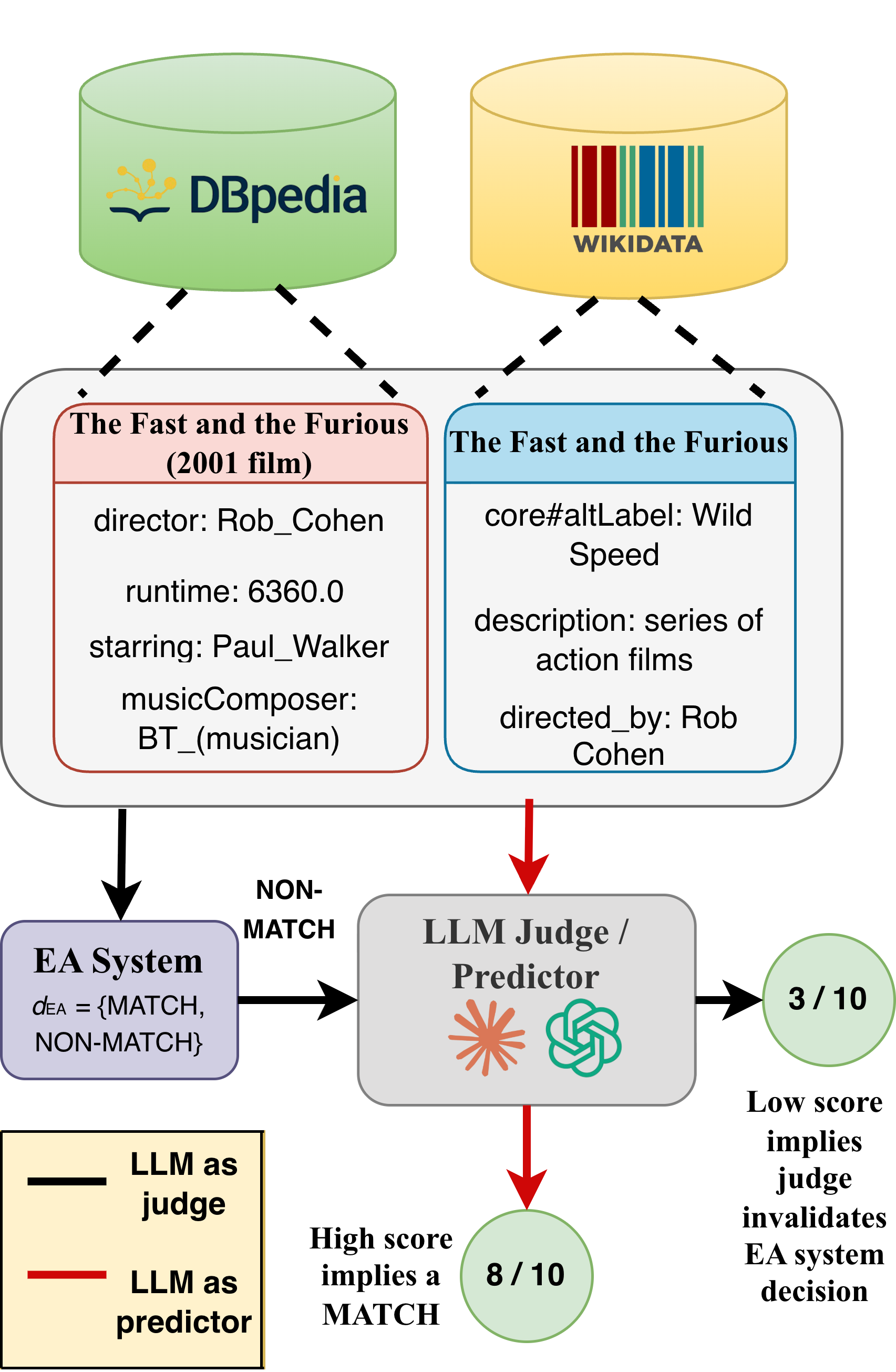}
\caption{LLM-as-judge evaluation for entity alignment, illustrated using D-W-15K (DBpedia--Wikidata). The two KG entities share the name ``The Fast and the Furious'' and the same director, but refer to different things: Entity~A (DBpedia) is the 2001 film; Entity~B (Wikidata) is the entire franchise series. The correct alignment label is \textsc{Non-Match}. \textbf{Black path (LLM as judge):} The EA system correctly predicts \textsc{Non-Match}; the LLM judge scores this decision 3/10, implying it considers \textsc{Non-Match} unjustified---anchoring on the shared name and director rather than reasoning from evidence. \textbf{Red path (LLM as predictor):} When the entity pair is sent directly to the LLM without any system label, it predicts \textsc{Match} with 8/10 confidence---again anchored on surface name similarity.
}
\label{fig:overview}
\end{figure}

For EA evaluation, this concern is acute: a judge must understand
entity semantics, reason about alignment plausibility independent of
system labels, and resist memorized entity names from pre-training.
When a system outputs a binary label (MATCH/NON MATCH) visible to the
judge, the judge may validate rather than evaluate, rationalizing the
system's decision post-hoc. Figure~\ref{fig:overview} illustrates this evaluation setting on a
D-W-15K example. The judge is not asked to predict alignment directly;
rather, it is asked to score whether the displayed EA decision is
justified by the evidence. This creates a structural vulnerability:
the decision label can become a cue to be rationalized rather than an
object to be evaluated. We refer to this failure mode as
\emph{anchor bias}: the judge's score may track the visible system
label even when the evidence should lead it to reject that label.
\textbf{Can we trust the judge?}

We present the first systematic benchmarking study of LLM-as-judge
reliability for entity alignment, introducing a three-experiment
framework of perturbation diagnostics, meta-evaluation across EA
systems, and counterfactual label-flip analysis
\cite{kaushik2020learning, gardner2020evaluating}. We also release
a new biomedical EA benchmark
(MeSH $\times$ SNOMED-CT, 15K pairs) \cite{ravideshik2026meshsnomed, biomed2026huggingface}.
Key contributions include:

\begin{itemize}[itemsep=2pt, topsep=2pt, parsep=0pt]
  \item We characterize \emph{anchor bias}, a failure mode
  in EA evaluation that is structurally distinct from prior
  anchoring and label-leakage effects observed in generation
  settings (\S\ref{sec:related}), and introduce a diagnostic suite (PBI,
  JSR~$\rho$, MRI) that combines perturbation and robustness
  measurement principles into a framework purpose-built for
  auditing structured-prediction LLM judges.

  \item We show that judge discrimination collapses when the system label
  is visible (J-ROC-AUC 0.12--0.87), while an independent
  alignment-prediction protocol recovers strongly positive alignment
  discrimination on distinctive-name datasets. Counterfactual label flips
  show that the visible label causally moves judge scores across both
  EasyEA \cite{cheng-etal-2025-easyea} and NeuSymEA
  \cite{chen2025neusymea} (FSR 53--99\%), with larger shifts for stronger
  judges; we treat this as label dependence of mixed direction, not proof
  of a single mechanism. A blinded two-annotator human
  evaluation provides a gold-label reference (\S\ref{sec:human-eval}).

  \item We construct and release a new biomedical EA benchmark
  (MeSH $\times$ SNOMED-CT, 15K pairs) with a reproducible
  auditing framework for EA-centric LLM judge reliability. 
  MeSH and SNOMED~CT use fundamentally different schema designs; hierarchical trees versus description-logic
  concepts, making alignment structurally harder and
  exposing failure modes that homogeneous benchmarks
  do not. The benchmark is released under the licensing terms
  described in \S\ref{sec:experiments} and Appendix~E.
\end{itemize}

\section{Related Work}\label{sec:related}

\paragraph{Entity Alignment.}
Entity alignment has progressed from translation-based embedding
methods \cite{bordes2013translating, sun2017cross, wang2018cross}
to LLM-augmented systems \cite{zhang2024making} and hybrid
neural-symbolic approaches such as EasyEA \cite{cheng-etal-2025-easyea}
and NeuSymEA \cite{chen2025neusymea}, with systematic benchmarking
across these approaches \cite{sun2020openea} driving
substantial progress. Closely related variants beyond NLP go by
the names of entity resolution and record linkage in
databases and information retrieval~\cite{getoor2013entity,
papadakis2020blocking}, and the problem also bears resemblance to
coreference resolution~\cite{ng2010supervised}, though EA assumes
entities are structured within a graph or database rather than
mentioned in free text. Despite these advances, large-scale
evaluation remains a bottleneck: verifying predicted alignments
requires expert annotation that is expensive and slow to produce.
\paragraph{LLM-as-Judge Evaluation and Its Failure Modes.}
LLM judges have been applied to text generation, summarization,
and preference evaluation \cite{zheng2023judging, fu2023gptscore},
offering scalability without gold-standard references. However,
documented failure modes position bias \cite{wang2023large},
verbosity bias \cite{dubois2024length}, and prompt sensitivity
\cite{webson2023do} raise systematic reliability concerns, and
none address structured
prediction tasks where the judge's input contains a discrete
decision label. Anchor bias builds directly on the established
anchoring-effects \cite{tversky1974judgment} and confirmation-bias
\cite{nickerson1998confirmation} literatures, and connects to
recent evidence that anchoring persists in frontier LLMs:
\citet{oleary2025anchoring} show that ChatGPT, Claude, and Gemini
all shift numerical judgments toward provided anchor values.
We do not claim anchoring itself as a new phenomenon;
our contribution is a structurally
distinct instantiation for structured prediction: prior work
studies anchoring on open-ended numerical estimates or opinion
questions where the anchor is contextual information, whereas here
the anchor is a discrete system decision label embedded directly in
the evaluation input, creating a rationalization target absent
from generation-evaluation settings, a distinction we return to
in \S\ref{sec:results}, where we show chain-of-thought prompting
does not resolve the bias.
\paragraph{LLM Reasoning Fragility and Shortcut Learning.}
LLMs are brittle to surface-level input changes even when
underlying semantics are preserved. GSM-Symbolic
\cite{mirzadeh2025gsmsymbolic} shows that a single irrelevant
clause causes performance drops of up to 65\% across frontier
models, suggesting that models replicate training patterns rather
than reason from first principles, connecting directly to shortcut learning
\cite{geirhos2020shortcut}: models rely on memorized surface correlations
rather than structural evidence, enabling judges to rationalize arbitrary system
decisions when entity names are recognizable. We build on this
to characterize anchor bias and report that, across the three frontier
models we test, label sensitivity grows rather than shrinks with model
scale, even though the same models are stronger label-free predictors; we
term this dissociation the \emph{frontier model paradox}
(\S\ref{sec:frontier-paradox}) and treat it as an observed pattern, not a
settled mechanism.
\paragraph{Counterfactual Evaluation.}
Counterfactual and contrast-set methods establish causal evidence
in NLP by holding content fixed while manipulating specific
factors \cite{kaushik2020learning, gardner2020evaluating}. We adapt
this methodology directly to EA judge evaluation:
flipping system decision labels while holding all entity evidence
fixed isolates the visible label as the only changed input, giving causal,
rather than correlational, evidence of label dependence; whether that
dependence is appropriate (checking the label against the evidence) or
biased is addressed through the direction of the shifts and
Experiments~2 and~4.
\paragraph{Knowledge Graph Benchmarking.}
Existing EA benchmarks \cite{sun2020openea} focus on homogeneous
general-domain graph pairs such as DBpedia--Wikidata and
DBpedia--YAGO, where source graphs share consistent schemas.
Biomedical knowledge graphs present a harder challenge due to
schema heterogeneity, sparse expert annotation, and
domain-specific terminology; we release MeSH--SNOMED~CT
\cite{ravideshik2026meshsnomed, biomed2026huggingface} as a large-scale biomedical
EA benchmark embedded in its full ontological context, directly
addressing this gap (licensing details in \S\ref{sec:experiments},
Appendix~E).

\section{Methodology}\label{sec:method}

\subsection{Entity Alignment Formulation}

Given two knowledge graphs $G_1$ and $G_2$, entity alignment
determines which pairs of entities $(e_1, e_2)$ refer to the same
real-world concept. An EA system $\mathcal{S}$ produces a decision
$d \in \{\text{MATCH}, \text{NON MATCH}\}$ for each candidate pair.
When human annotation is unavailable, an LLM judge $\mathcal{J}$
evaluates whether this decision is correct by assigning a score
$s \in [0,10]$. The central question is whether $s$ genuinely
reflects decision correctness or is confounded by the visible label
$d$ itself. If the judge simply rationalizes whatever label it sees,
LLM-as-judge evaluation is circular: it reproduces the system's
decisions rather than auditing them.

\subsection{Anchor Bias: Definition and Formalization}

We define \emph{anchor bias} as the tendency of an LLM judge to
rationalize a displayed system decision label rather than
independently assess the evidence for entity alignment. When a
judge receives the tuple $(e_1, e_2, \text{context}, d)$, anchor
bias occurs when $s$ is driven primarily by $d$ rather than by the
entity evidence. This makes judge scores uninformative: a judge
that validates system labels appears to perform well on correct
predictions and poorly on incorrect ones not because it is
reasoning, but because it is echoing. The judge becomes a
validator rather than an independent arbiter, defeating the
purpose of automated evaluation.

\subsection{Perturbation Bias Diagnostics}

To claim that judges rationalize visible labels rather than
reasoning from evidence, we must first establish what judges
rely on in the \emph{absence} of a label. If scores collapse
when entity names are anonymized, the judge's baseline reasoning
is driven by memorized name recognition; precisely the surface
shortcut that makes it susceptible to label anchoring. We administer controlled
modifications to entity names, evidence, and context following
the behavioral testing paradigm \cite{ribeiro2016model}, and
characterize judge dependence through three complementary metrics:

\paragraph{Perturbation Bias Index (PBI).} PBI measures the average score drop after perturbation:
\begin{equation}
  \text{PBI} = \bar{s}_{\text{original}} - \bar{s}_{\text{perturbed}}
\end{equation}
A value $> 0$ indicates reliance on the perturbed feature rather
than structural evidence.

\paragraph{Judge Score Rank Correlation (JSR~$\rho$).} JSR is the
Spearman rank correlation between original and perturbed scores:
\begin{equation}
  \text{JSR} = \rho\bigl(\mathbf{s}_{\text{original}},\,
  \mathbf{s}_{\text{perturbed}}\bigr)
\end{equation}
High JSR indicates consistent reliance on the perturbed feature
across pairs; low JSR indicates erratic, pair-specific scrambling
that PBI alone cannot reveal.

\paragraph{Massive Rank Inversion Rate (MRI).} MRI measures the
fraction of pairs with score collapse exceeding 2 points:
\begin{equation}
  \text{MRI} = \frac{|\{i : s^i_{\text{original}} -
  s^i_{\text{perturbed}} > 2\}|}{n} \times 100\%
\end{equation}
PBI can stay small even when a minority of pairs collapse
catastrophically; MRI isolates these extreme failures. Together,
PBI, JSR, and MRI profile name-dependence magnitude, consistency,
and catastrophic rate.

\subsection{EA System Meta-Evaluation}

Perturbation diagnostics reveal what the judge relies on, but
not whether it can distinguish correct from incorrect EA
predictions; the operational question for evaluation. To measure
this, and to isolate the role of label exposure, we contrast
three prompting conditions that vary the structural relationship
between the judge and the system decision: (I)~\emph{Evidence-locked
decision validation} 8-step chain-of-thought reasoning
instructing the judge to assess the system decision from entity
evidence alone; (II)~\emph{Minimal decision validation} no
guidance, representing the default practitioner setting;
(III)~\emph{Label-free alignment prediction} the system
decision is never shown; the LLM independently predicts
alignment. Prompts~I--II are LLM-as-judge conditions;
Prompt~III is an LLM-as-predictor capability probe.

We operationalize judge performance through the metrics in Table~\ref{tab:ea_metrics}.

\begin{table}[t]
\centering
\small
\setlength{\tabcolsep}{4pt}
\caption{Metrics for EA system meta-evaluation (Experiment~2).
Under Prompts~I--II the LLM scores how well-justified the visible
system decision is; J-ROC-AUC measures correctness discrimination.
Under Prompt~III the LLM acts as a predictor; no label is shown
and Score Gap is the primary metric.}
\label{tab:ea_metrics}
\begin{tabular}{p{1.2cm}p{2cm}p{2.2cm}}
\toprule
\textbf{Metric} & \textbf{Formula} & \textbf{Interpretation} \\
\midrule
Score Gap & $\Delta = \bar{s}_{\mathrm{M}} - \bar{s}_{\mathrm{NM}}$ & Negative $\Delta$ = inverted discrimination \\
PDV & $\mathrm{Var}(s \mid \text{prompt})$ & Low PDV = compression \\
J-ROC-AUC & $P(s_{+} > s_{-})$ & ${<}0.5$ = anchor bias (P-I/II, judge); ${>}0.9$ = strong alignment prediction (P-III, predictor) \\
\bottomrule
\end{tabular}
\end{table}

\subsection{Counterfactual Label Sensitivity}

Meta-evaluation establishes whether judges discriminate correctly,
but a correlation between label and score does not establish
causation. Establishing that labels \emph{cause} score changes requires an
intervention: holding all entity evidence fixed and changing
only the visible label \cite{kaushik2020learning}. Any score
change under this intervention is attributable solely to the label.
Because Prompts~I and~II ask the judge to score whether the displayed
decision is correct, some label dependence is required of a good judge, so
the experiment tests how strongly, and in which direction, scores respond,
not whether they respond. We quantify the response through three metrics:

\paragraph{Mean Absolute Shift (MAS).} MAS measures the average magnitude of score change after label flipping:
\begin{equation}
  \text{MAS} = \frac{1}{n} \sum_{i=1}^{n}
  \left| s^{(i)} - \bar{s}^{(i)} \right|
\end{equation}
High MAS indicates that scoring depends strongly on the visible
label.

\paragraph{Flip Sensitivity Rate (FSR).} FSR measures the fraction of pairs with score shifts exceeding 2 points after label flipping:
\begin{equation}
  \text{FSR} = \frac{|\{i : |s^{(i)} - \bar{s}^{(i)}| > 2\}|}
  {n} \times 100\%
\end{equation}
While MAS captures average sensitivity, FSR captures the
\emph{prevalence} of large causal effects: high FSR means label
dependence is systematic rather than marginal, affecting the
majority of evaluation decisions. FSR alone cannot separate appropriate sensitivity from bias.

\paragraph{Label Following Rate (LFR).} LFR measures the fraction of pairs where the score shifts in the direction of the flipped label:
\begin{equation}
\begin{split}
\text{LFR} = \frac{100}{n} \Bigl(
&\sum_{i \in \mathcal{M}} \mathbb{1}[\bar{s}^{(i)} < s^{(i)}] \\
&+ \sum_{i \in \mathcal{N}} \mathbb{1}[\bar{s}^{(i)} > s^{(i)}]
\Bigr)\%
\end{split}
\end{equation}
Downward when \textsc{Match} flips to \textsc{Non-Match}, upward
when \textsc{Non-Match} flips to \textsc{Match}.
LFR is a directional descriptor, not a test of bias. Because the scores
rate the correctness of the displayed decision, an evidence-sensitive judge
should lower its score whenever a correct decision is flipped to an incorrect
one. For \textsc{Non-Match} pairs this score moves \emph{down}, against the
direction LFR counts, so such a judge yields LFR of at most about 50\% (the
exact value depends on how many original decisions were correct). LFR clearly
above 50\% indicates scores moving toward the displayed label even when it
contradicts the evidence. FSR captures whether scores respond to a flip, LFR
how often the response follows the new label; we read neither alone as
evidence of anchoring.

\subsection{Human Evaluation Protocol}\label{sec:human-eval-method}

To validate that judge scores reflect genuine EA correctness rather
than an artifact of our automated metrics, we conducted a blinded
two-annotator human evaluation: both authors independently scored
the same pairs across four negative-sampling strategies under all
three prompts, without access to each other's judgments during
annotation. Full sampling details, inter-annotator agreement
statistics, and gold-label accuracy are reported in
\S\ref{sec:human-eval}.

\section{Experimental Setup}\label{sec:experiments}
\paragraph{Datasets.}
We evaluate on three datasets spanning distinct alignment
challenges. \textbf{DW} (DBpedia $\times$ Wikidata) and \textbf{DY}
(DBpedia $\times$ YAGO) are OpenEA benchmarks \cite{sun2020openea}
with distinctive and near-identical entity names respectively,
providing a controlled contrast between name-rich and name-ambiguous
conditions. \textbf{BIO} (MeSH $\times$ SNOMED-CT, 15K pairs,
publicly released \cite{ravideshik2026meshsnomed, biomed2026huggingface}) is the first
large-scale biomedical EA benchmark released with full ontological
context and is a key contribution of this work.

\paragraph{BIO Benchmark Construction.}
Ground-truth alignments were extracted from UMLS cross-references
between MeSH (December~2023) and SNOMED~CT (March~2026), yielding
46,823 raw MeSH--SNOMED~CT alignment pairs from which 15,000 were
stratified-sampled. Knowledge graph triples were extracted directly
from source dumps: MeSH contributes 11.4M attribute and 6.9M
relation triples; SNOMED~CT contributes 1.4M attribute and 1.3M
relation triples.

BIO presents three structural challenges absent from general-domain
benchmarks: no shared predicate vocabulary between MeSH and
SNOMED~CT, sparse LLM pre-training signal over biomedical ontology
concepts, and genuine many-to-many alignments from SNOMED~CT's
finer granularity. Full construction details, dataset statistics,
and release information are provided in Appendix~E
\cite{ravideshik2026meshsnomed, biomed2026huggingface}.

\paragraph{Licensing and Access.}
BIO is derived from two source ontologies with distinct access
terms. MeSH is distributed by the U.S. National Library of Medicine
(NLM) under its standard terms of use, which permit research and
redistribution with attribution. SNOMED~CT requires a UMLS
Metathesaurus license, free for researchers and institutions in
UMLS-affiliate countries, obtained directly from the U.S. National
Library of Medicine; redistribution of SNOMED~CT-derived content is
permitted only to other UMLS license holders. We release BIO's
triples and gold alignment pairs under these same conditions:
users must independently hold a valid UMLS license to access the
SNOMED~CT-derived portion of the benchmark, consistent with the
original providers' terms (full details in Appendix~E).

\paragraph{EA Systems.}
We evaluate two EA systems and their ablated variants spanning diverse architectures to test
whether observed biases are judge-driven rather than
system-specific:
\begin{enumerate}
  \item \textbf{EasyEA}  \cite{cheng-etal-2025-easyea} (Hits@1:
0.806/0.947/0.518 on DW/DY/BIO) combines embedding-based
representations with LLM guidance; \textit{EasyEA-EmbOnly} is its
embedding-only ablation. 
\item \textbf{NeuSymEA} \cite{chen2025neusymea}
(Hits@1: $\approx$0.803/0.973 on DW/DY) integrates neural and
symbolic reasoning via variational inference;
\textit{NeuSymEA-NoRefine} is its base variant without symbolic
refinement.
\end{enumerate}

\paragraph{EA System Output and Sampling Protocol.}
EasyEA~\cite{cheng-etal-2025-easyea} produces alignments via a three-stage
pipeline combining LLM-based summarization, embedding fusion, and
hierarchical LLM candidate selection; EasyEA-EmbOnly disables the
Stage~3 LLM selection module. NeuSymEA~\cite{chen2025neusymea}
integrates symbolic reasoning with a neural embedding model via
variational EM; NeuSymEA-NoRefine disables the symbolic refinement
component.

For Experiments~2 and~3, pairs were sampled directly from EA system
output without consulting ground-truth labels. 250 \textsc{Match}
pairs were sampled uniformly at random from the EA system's predicted
alignments. 250 \textsc{Non-Match} pairs were obtained by randomly
sampling entity pairs $(e_1, e_2)$ with $e_1 \in G_1$, $e_2 \in G_2$,
rejecting any pair present in the EA system's predicted alignments.
Since EA predictions constitute a negligible fraction of
$G_1 \times G_2$, the rejection rate is negligible. Ground-truth
labels are used only when computing evaluation metrics after sampling,
not during pair selection. System outputs were fixed before any judge
evaluation began. Full details of the sampling procedure for all
experiments are in Appendix~I. For Experiment~4 (human evaluation),
we additionally sample \textsc{Non-Match} pairs from three harder
negative-mining strategies (lexical-neighbor, TF-IDF nearest-neighbor,
and high-confidence false positives), described in
\S\ref{sec:human-eval}.

\paragraph{LLM Judges.}
We evaluate three frontier models as judges: \textbf{GPT-4o-mini},
\textbf{GPT-4o}, and \textbf{Claude Opus~4.7}, accessed via
OpenRouter with temperature~$= 0$ and
\texttt{max\_completion\_tokens}~$= 1500$ across all conditions.
All calls were routed directly to each provider's official endpoint
(OpenRouter metadata confirms \texttt{provider: "Anthropic"} for
Claude Opus~4.7 calls, resolving consistently to
\texttt{anthropic/claude-opus-4-7}, and OpenAI's official endpoint
for GPT-4o and GPT-4o-mini), and all calls within a given experiment
were completed within a single continuous session to minimize the
risk of snapshot drift. We note that temperature~$=0$ does not by
itself formally guarantee bit-for-bit determinism across arbitrary
model snapshots; we are not aware of any snapshot changes for these
models during our experimental window, and we release full
intermediate logs to support replication (Appendix~C).
Total API expenditure was \$577.17 across all experiments
(153,000 calls, 238.8M tokens); a full cost breakdown is provided
in Appendix~C.
\subsection{Experiment 1: Perturbation Bias Diagnostics}
We conduct perturbation tests over $n{=}500$ pairs per condition
across all three datasets and all judge models. Pairs are sampled
uniformly at random from ground-truth alignments (MATCH pairs only;
no NON-MATCH class in Experiment~1). We use seven
perturbation strategies (P1--P7) described in Appendix~A: identity
anonymization (P1), evidence suppression at 30\% and 50\% (P2,
P3), surface transformation (P4), order randomization (P5),
identifier removal (P6), and structured-to-prose conversion (P7).
P1 is the most diagnostic, directly ablating the name recognition
shortcut; P2, P3, and P6 are reported alongside P1 in
Table~\ref{tab:perturbation} to capture evidence suppression
dose-response and identifier dependence; P4, P5, and P7 are
reported in Appendix~G (Table~\ref{tab:perturbation_remaining}),
with P5 serving as the negative control. For each condition we compute PBI, JSR, and MRI to
characterize the judge's feature dependence profile. Sample size
justification ($n{=}500$) is provided in Appendix~D.0.
\subsection{Experiment 2: EA System Meta-Evaluation}
We evaluate all combinations of datasets (DW, DY, BIO), EA
systems (EasyEA and its ablation EasyEA-EmbOnly, NeuSymEA and its ablation NeuSymEA-NoRefine),
judges (GPT-4o-mini, GPT-4o, Claude Opus~4.7), and prompts
(I, II, III). For each condition we sample $n{=}500$ pairs
(250 MATCH, 250 NON MATCH) and compute Score Gap, PDV, and
J-ROC-AUC.

Under Prompts~I--II, J-ROC-AUC measures
$P(s_{\text{correct decision}} > s_{\text{incorrect decision}})$;
inversion below 0.5 is the operational signature of anchor bias.
Under Prompt~III, Score Gap is the primary metric; J-ROC-AUC
is also reported
but measures alignment prediction capability
$P(s_{\textsc{Match}} > s_{\textsc{Non-Match}})$, not
correctness discrimination. Full threshold sweeps are in Appendix~B
(Tables~\ref{tab:thresh_easy_dw}--\ref{tab:thresh_neu}).

\subsection{Experiment 3: Counterfactual Label Sensitivity}
We test causal label effects by flipping system decisions while
holding entity evidence fixed, sampling $n{=}100$ pairs per
condition (50 MATCH $+$ 50 NON MATCH) from EasyEA
\cite{cheng-etal-2025-easyea} and NeuSymEA \cite{chen2025neusymea} on DW
and DY under Prompts~I and~II. BIO is excluded as NeuSymEA was
not evaluated on that dataset; Prompt~III is excluded as it
exposes no label. For each pair we obtain scores under the
original and flipped decision and compute MAS, FSR, and LFR to
quantify how strongly, and in which direction, the visible label moves
the score; we treat this as a causal sensitivity analysis, not a
stand-alone test of bias.

\subsection{Experiment 4: Human Evaluation}\label{sec:human-eval}

To validate that anchor bias reflects genuine judge unreliability
rather than an artifact of our automated diagnostics, we conducted
a blinded two-annotator human evaluation following standard
annotation protocols (\S\ref{sec:human-eval-method}). Both authors
independently scored the same 102 pairs (50 \textsc{Match}, 52
\textsc{Non-Match} spanning all four negative-sampling strategies)
under Prompts~I, II, and III, for 306 scored instances per
annotator, without seeing each other's scores during annotation.

\paragraph{Inter-annotator agreement.} Agreement between the two
annotators is strong: Spearman $\rho = 0.967$, Pearson $r = 0.985$,
Cohen's $\kappa = 0.902$, and MAE $= 0.38$, indicating almost
perfect agreement \cite{landis1977measurement}. Agreement holds
separately for \textsc{Match} pairs ($\kappa = 0.731$) and
\textsc{Non-Match} pairs ($\kappa = 0.891$).

\paragraph{Gold-label accuracy.} The consensus annotation (mean of
the two annotators' scores) achieves a Score Gap of 7.896 points
between \textsc{Match} and \textsc{Non-Match} pairs (9.800 vs.\
1.904), with 100\% of \textsc{Match} pairs correctly scored high
and 82.7\% of \textsc{Non-Match} pairs correctly scored low,
yielding an overall gold-label accuracy of 91.3\%. This confirms
the human annotations are both internally consistent (via
inter-annotator agreement) and externally valid (via agreement
with known ground truth).

\paragraph{Human--LLM correlation and the anchor-bias mechanism.}
Using the consensus annotation as the gold standard, all three LLM
judges show strong overall correlation with human judgment:
GPT-4o-mini ($\rho = 0.860$, $\kappa = 0.743$), GPT-4o
($\rho = 0.893$, $\kappa = 0.812$), and Claude Opus~4.7
($\rho = 0.920$, $\kappa = 0.871$, MAE $= 0.23$). Claude Opus~4.7
achieves the highest alignment with human consensus, consistent
with its status as the strongest of the three frontier judges we
evaluate. The per-label breakdown adds detail beyond the aggregate:
under label-exposing prompts, \textsc{Match} pairs show low
human--LLM correlation ($\rho = 0.228$ to $0.412$) while
\textsc{Non-Match} pairs show high correlation ($\rho = 0.779$ to
$0.867$). This asymmetry is compatible with anchor bias but does not establish it:
human scores on \textsc{Match} pairs have little spread (100\% were scored
high), which alone lowers a rank correlation, and the per-label subsets are
small ($n{=}50$ and $n{=}52$).
Table~\ref{tab:human_eval} summarizes all of these
human evaluation statistics together.

\section{Results}\label{sec:results}

\subsection{Perturbation Bias Diagnostics}

\textbf{Identity anonymization is the singular catastrophic
perturbation trigger on BIO (MRI up to 69\%), while general-domain
judges remain stable across all conditions.}

\begin{figure}[t]
\centering
\includegraphics[width=0.95\columnwidth, height=0.29\textheight, keepaspectratio]{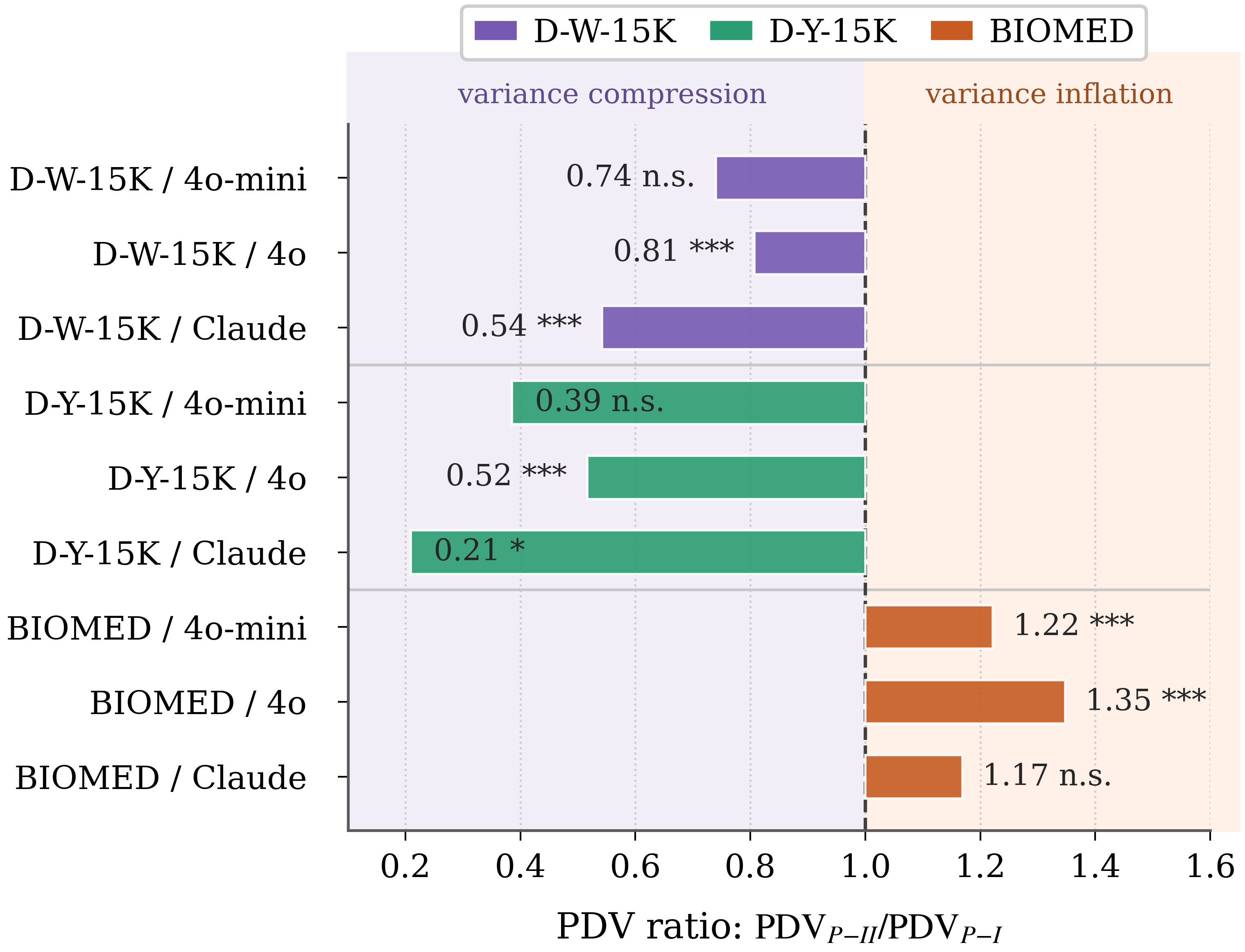}
\caption{PDV ratio ($\text{PDV}_{P\text{-}II}/\text{PDV}_{P\text{-}I}$)
per dataset--judge condition. Ratios ${<}1$ indicate variance compression
under minimal prompting; ratios ${>}1$ indicate inflation.
General-domain judges compress uniformly (0.21--0.81);
BIO judges inflate (1.17--1.35), reflecting absent pre-training signal.
Significance: $^{***}p{<}0.001$, $^{*}p{<}0.05$, n.s.\ $p{\geq}0.05$.}
\label{fig:pdv}
\end{figure}

Minimal prompting (Prompt~II) elicits systematically higher baseline
scores across all nine conditions. As shown in Figure~\ref{fig:pdv}
(full statistics in Table~\ref{tab:pdv}, Appendix~H), for
general-domain datasets (DW, DY), PDV ratios fall below 1.0
(range: 0.21--0.81), indicating that removing prompt guidance
compresses score variance: the judge collapses onto memorized entity
knowledge and assigns uniformly high scores regardless of pair quality.
BIO ratios exceed 1.0 (range: 1.17--1.35), reflecting absent
pre-training signal; full analysis in \S\ref{sec:results}.4.

As shown in Table~\ref{tab:perturbation}, P1 (identity
anonymization) induces MRI of 60.6--68.8\% on BIO in six of nine judge--prompt
conditions (18.4--48.0\% in the other three). No other
perturbation strategy exceeds 7\% MRI on BIO, and P5 (order
randomization), serving as a negative control, produces MRI
of at most 8.0\% (below 5\% in 25 of 27 conditions), confirming the framework
identifies genuine name-dependence rather than noise.

\subsection{EA System Meta-Evaluation}

\textbf{Anchor bias is endemic: under label-exposing prompts J-ROC-AUC is
0.14--0.53 on DW and BIO (0.12--0.87 across all datasets), while Prompt~III (LLM as predictor) recovers near-ceiling discrimination.}

Score Gaps are predominantly negative under both label-exposing
prompts (Appendix~F, Table~\ref{tab:scoregap_sig}), confirming
that judges assign systematically higher validation scores to
\textsc{Non-Match} decisions than to \textsc{Match}
decisions---the behavioral signature of anchor bias. Structured
chain-of-thought reasoning (Prompt~I) does not mitigate the bias;
in several conditions it amplifies it.

Across DW and DY, 18 of 24 conditions achieve statistically
significant positive discrimination under Prompt~III (LLM as predictor)
($p{<}0.05$, Table~\ref{tab:scoregap_sig}); the remaining 6 conditions
are concentrated in DY, where near-identical entity names limit
label-free discrimination. The capability
gap is architecture-dependent: EasyEA~\cite{cheng-etal-2025-easyea}
variants collapse under Prompt~III on DY, where near-identical
entity names provide no discriminative signal in the absence
of a label, while NeuSymEA~\cite{chen2025neusymea} variants
achieve J-ROC-AUC 0.981--1.000 and Score Gap $+$7.9--$+$9.1
($p{<}0.001$) on both datasets, indicating that structure-aware
alignment training confers robustness to predictor-mode evaluation.

Label-free evaluation succeeds on distinctive-name datasets and
structure-aware systems but fails on near-identical names paired
with weak EA systems; full scope analysis in Appendix~J.

\subsection{Counterfactual Label Sensitivity and the Frontier Model Paradox}\label{sec:frontier-paradox}

\textbf{Visible labels causally shift judge scores by more than 2 points
in 53--99\% of pairs, and the shifts grow with model capability: stronger
judges are more label-sensitive, not less.}

As shown in Table~\ref{tab:labelflip}, FSR exceeds 50\% in every condition
under Prompt~I. Prompt~II raises label sensitivity further
(Appendix~D.3, Table~\ref{tab:labelflip_p2}): MAS is higher in all 15
conditions and FSR in 13 of 15. Under Prompt~I, MAS and FSR rise from
GPT-4o-mini to GPT-4o to Claude Opus~4.7 in every condition, with Claude
reaching FSR~$=$~99\% on DW under EasyEA and 95\% under NeuSymEA; under
Prompt~II, Claude has the highest MAS in every condition and FSR is near
ceiling (76--100\%).

LFR is mixed. Under Prompt~I it is at or below 50\% in 12 of 15 conditions
(exceptions: DY with GPT-4o 53\% and Claude 65\% under EasyEA, BIO with
Claude 51\%); under Prompt~II it ranges from 37\% to 80\%, with DY under
EasyEA the only setting where all three judges are well above 50\%
(70--80\%). As noted in \S\ref{sec:method}, an evidence-sensitive validator
is itself expected to yield LFR at or below about 50\%, so values near or
below 50\% do not show that scores follow the label rather than the
evidence. The DY--EasyEA values are the clearest sign of scores moving
toward the displayed label, possibly reflecting the near-identical names there.

\begin{figure}[t]
\centering
\includegraphics[width=0.95\columnwidth, height=0.36\textheight, keepaspectratio]{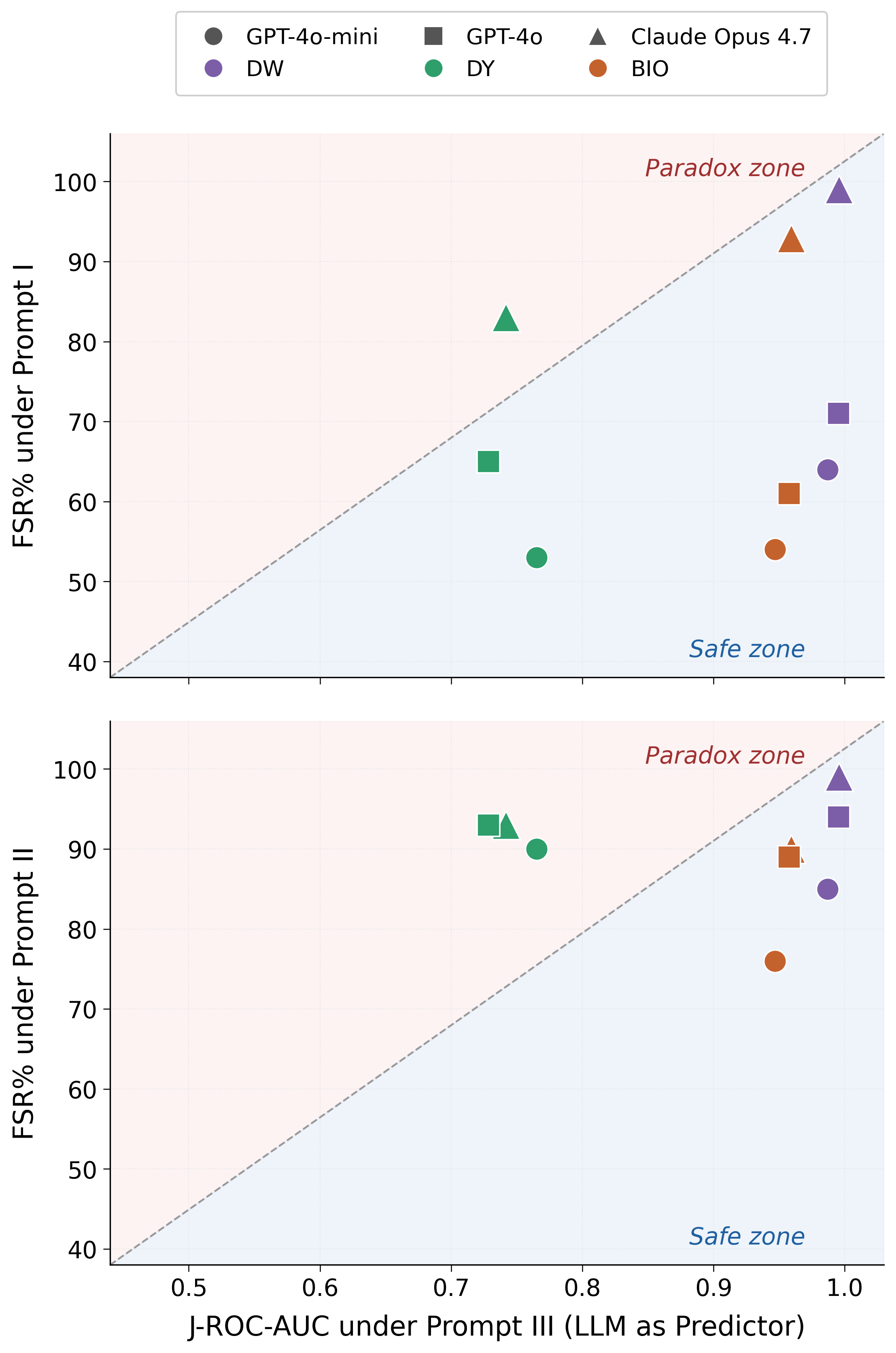}
\caption{Frontier model paradox: FSR\% (label sensitivity under
Prompts~I and~II, LLM as judge) vs.\ J-ROC-AUC (alignment prediction under
Prompt~III, LLM as predictor). Stronger judges show greater label
sensitivity and are also more capable when labels are withheld; the figure
describes this pattern and does not establish its cause.
Shape = judge; color = dataset.}
\label{fig:paradox}
\end{figure}

We read FSR as showing that the visible label is a strong causal input to
the score, as it must be for a validator, and the mixed LFR as showing that
this dependence is not uniformly label-following. The growth of
sensitivity with model capability is the frontier model paradox; we report it
as an observed pattern, not as proof that stronger judges are more biased,
since larger shifts could also reflect stronger judges detecting flipped
decisions more reliably. Figure~\ref{fig:paradox} relates FSR to label-free
J-ROC-AUC; that comparison also changes the task (\S\ref{sec:limitations}).

\paragraph{Mechanism and generalization.} A candidate mechanism (pre-training
signal density) and the expected extension to other structured prediction
tasks are discussed in Appendix~J; both are untested hypotheses.

\begin{table}[t]
\centering
\tiny
\setlength{\tabcolsep}{2pt}
\renewcommand{\arraystretch}{0.70}
\caption{Counterfactual label-flip results under Prompt~I: MAS,
FSR\%, 95\% CI, and LFR\% ($n{=}100$ per condition, 50 MATCH $+$ 50 NON MATCH).
FSR exceeds 50\% in every condition. 95\% bootstrap CIs ($B{=}10{,}000$) for FSR have lower bounds above 50\%
in most conditions; the exceptions are GPT-4o-mini on DY (EasyEA [43, 63],
NeuSymEA [47, 66]) and on BIO ([44, 64]).
Compare to Prompt~II amplification in
Table~\ref{tab:labelflip_p2}.}
\label{tab:labelflip}
\resizebox{\columnwidth}{!}{%
\begin{tabular}{llcccc}
\toprule
\textbf{Dataset--Judge} & \textbf{System} &
\textbf{MAS} & \textbf{FSR\%} & \textbf{95\% CI} & \textbf{LFR\%} \\
\midrule
DW 4m & EasyEA   & 3.55 & 64 & [55, 73] & 17 \\
DW 4o & EasyEA   & 4.66 & 71 & [62, 80] & 29 \\
DW O4 & EasyEA   & 6.07 & 99 & [97, 100] & 50 \\
DW 4m & NeuSymEA & 4.34 & 62 & [52, 71] & 41 \\
DW 4o & NeuSymEA & 6.39 & 81 & [73, 88] & 46 \\
DW O4 & NeuSymEA & 8.21 & 95 & [90, 99] & 47 \\
\midrule
DY 4m & EasyEA   & 3.71 & 53 & [43, 63] & 43 \\
DY 4o & EasyEA   & 4.80 & 65 & [56, 74] & 53 \\
DY O4 & EasyEA   & 6.27 & 83 & [75, 90] & 65 \\
DY 4m & NeuSymEA & 4.34 & 57 & [47, 66] & 39 \\
DY 4o & NeuSymEA & 6.76 & 90 & [84, 95] & 48 \\
DY O4 & NeuSymEA & 7.97 & 94 & [89, 98] & 48 \\
\midrule
BIO 4m & EasyEA  & 3.26 & 54 & [44, 64] & 32 \\
BIO 4o & EasyEA  & 4.17 & 61 & [51, 70] & 35 \\
BIO O4 & EasyEA  & 5.88 & 92 & [86, 97] & 51 \\
\bottomrule
\end{tabular}}
\end{table}

\subsection{Biomedical Domain: A Distinct Failure Mode}

\textbf{BIO exhibits a structurally distinct failure mode:
perturbation collapse is pre-training-driven, label sensitivity is
somewhat smaller, and label-free recovery is weaker than on
general-domain datasets.}

On BIO, P1 identity anonymization collapse is prompt-invariant:
JSR turns negative for 4o-mini ($-$0.076, $-$0.142) and near-zero
for 4o (0.029, $-$0.026) under both Prompts~II and~III regardless
of label visibility, ruling out label anchoring as the primary
driver. Instead, MeSH and SNOMED~CT concepts are rarely encountered
verbatim in pre-training corpora, leaving judges without recognizable
anchors and producing higher-variance, evidence-driven scoring.

Label sensitivity is present but somewhat smaller: FSR reaches
54--92\% (vs.\ 62--99\% on DW), and LFR on BIO (32--51\%) lies within
the range seen on DW and DY (17--65\%). Label-free recovery
under Prompt~III is achievable---EasyEA achieves significant positive
discrimination across all three judges ($p{<}0.05$,
Table~\ref{tab:scoregap_sig}), with J-ROC-AUC 0.931--0.948---though
this falls short of near-ceiling recovery on DW (0.981--1.000),
reflecting sparse attribute overlap and heterogeneous relation
predicates in MeSH--SNOMED~CT.

\section{Conclusion}\label{sec:conclusion}
Judge failure in our experiments is not a simple reasoning deficit: it is
tied to the visible label and to entity name cues. Even under an 8-step
evidence-locked chain-of-thought protocol, LLM judges fail
to discriminate correct from incorrect EA predictions (J-ROC-AUC
0.12--0.87 overall; 0.14--0.53 on DW). Stronger
models are not automatically the solution: greater capability is associated
with greater label sensitivity, not less, although human agreement also rises
with model strength; we report this as an open pattern, not as evidence that
stronger judges are more biased. Prompt~III, which repurposes the LLM as a
predictor rather than a judge, reveals strongly positive alignment
discrimination when labels are withheld, though it also changes the task.
Counterfactual label flips shift scores by more than 2 points in 53--99\% of
pairs (FSR), with mixed direction (LFR 17--80\%); we read this as strong label
dependence, not on its own as anchoring. A blinded human evaluation provides a
gold-label reference. Practitioners should audit
LLM judges before relying on them, and consider label-free prediction
protocols where the system decision is not essential.
We release the BIO benchmark and all evaluation code at \url{https://github.com/vaibhavalakshmiravideshik/llm-as-a-judge-entity-alignment} to support this work.

\section*{Ethical Considerations}
This work evaluates LLM judge reliability using publicly available entity alignment benchmarks. No new datasets containing personal or sensitive information were collected. Our evaluation focuses on system behavior and does not involve human subjects. The BIO benchmark draws exclusively from public medical ontologies (MeSH and SNOMED-CT) that are freely available for research, subject to the licensing terms described in \S\ref{sec:experiments} and Appendix~E. We release our evaluation code to promote reproducible auditing of LLM-as-judge approaches. As LLM-based evaluation becomes more prevalent in AI research, systematic analysis of judge reliability in structured tasks is important for ensuring research integrity and preventing propagation of biased judgments in downstream applications.
\section*{Acknowledgments}
We use an AI assistant for paraphrasing support during the
writing of this manuscript.

\section*{Limitations}\label{sec:limitations}
We use 50/50 class-balanced evaluation ($n{=}500$) to isolate discrimination capability; absolute values should be interpreted relative to this protocol rather than operational class imbalance. We evaluate only three closed-source frontier judges; open-source models and non-English datasets remain untested. Prompt~III changes both label exposure and task framing
simultaneously; improvements cannot be attributed solely
to label removal without a matched control. Because Prompts~I and~II ask
the judge to score the correctness of the displayed decision, a large score
shift under label flipping is expected of a good judge; Experiment~3
therefore establishes causal label dependence, not bias, and the direction
of the shifts is mixed. The frontier
model paradox is observed across three frontier models
accessed via a single API provider; generalization to
open-source models or different model families remains
an open question, as does empirical validation of anchor bias
on structured prediction tasks beyond entity alignment
(\S\ref{sec:frontier-paradox}). Our human evaluation
(\S\ref{sec:human-eval}) uses two author-annotators rather than
independent, uninvolved expert raters; while blinding and strong
inter-annotator agreement mitigate this concern, an independent
expert annotation study would further strengthen the validation.

\vspace{-8pt}
\bibliography{custom}

\clearpage
\appendix
\raggedbottom
\setlength{\floatsep}{4pt plus 2pt minus 1pt}
\setlength{\textfloatsep}{6pt plus 2pt minus 2pt}
\setlength{\intextsep}{4pt plus 2pt minus 1pt}
\setlength{\parskip}{0pt}

\section*{Appendix Table of Contents}
\begin{tabular}{@{}p{2.8cm}@{}p{5.5cm}@{}}
\textbf{Appendix A} & Judge Prompt Templates \\
\quad A.1 & Prompt I: Evidence-Locked Evaluation \\
\quad A.2 & Prompt II: Minimal Evaluation \\
\quad A.3 & Prompt III: Independent Alignment Prediction \\
\textbf{Appendix B} & Judge Meta-Evaluation Threshold Sweeps \\
\textbf{Appendix C} & Computational Cost and API Usage \\
\textbf{Appendix D} & Robustness and Statistical Validation \\
\quad D.0 & Sample Size Justification \\
\quad D.1 & MRI Threshold Sensitivity \\
\quad D.2 & Bootstrap Confidence Intervals for PBI and MRI \\
\quad D.3 & Label-Flip Effects Under Prompt II \\
\textbf{Appendix E} & BIOMED Dataset Construction and Validation \\
\textbf{Appendix F} & Score Gap Analysis and Significance \\
\textbf{Appendix G} & Full Perturbation Bias Diagnostics \\
\textbf{Appendix H} & PDV Baseline Statistics \\
\textbf{Appendix I} & Sampling Procedures for All Experiments \\
\textbf{Appendix J} & Extended Discussion \\
\end{tabular}

\clearpage

\section*{Appendix A: Judge Prompt Templates}
\label{app:prompts}
We provide the complete prompt templates used in Experiments~1 and~2.
All three prompts share identical input structure (entity IDs, labels,
types, properties, relationships, and system decision where applicable)
and identical output parsing logic. They differ in task framing,
guidance level, and whether the system decision is exposed to the judge.

\subsection*{A.1 Prompt I: Evidence-Locked Evaluation}
Prompt~I is the most carefully engineered prompt in our study. It
instantiates a \emph{decision validation} task: the judge receives the
system decision and is explicitly instructed to suppress pre-training
knowledge and reason exclusively from the provided evidence through an
8-step chain-of-thought protocol.

\begin{lstlisting}[breaklines=true, breakatwhitespace=true, basicstyle=\ttfamily\footnotesize, frame=single, framesep=5pt]
SYSTEM ROLE
You are an Evidence-Locked Evaluator. Your sole function is to assess
whether an Entity Alignment system made a reasonable decision given
the evidence provided.

HARD CONSTRAINTS - READ BEFORE PROCEEDING:
1. EVIDENCE BOUNDARY: You must reason EXCLUSIVELY from the evidence
   strings provided below. No exceptions.
2. MEMORY SUPPRESSION: You are FORBIDDEN from using any external world
   knowledge, memorized entity facts, or prior training knowledge about
   these entities. If you recognize an entity, DISCARD that recognition.
   Treat every entity as if it were fictional and previously unknown
   to you.
3. REASONING REQUIREMENT: Every conclusion you draw must be traceable
   to a specific piece of evidence in the blocks below. If you cannot
   cite it, you cannot claim it.
4. VIOLATION CHECK: Before writing each conclusion, ask yourself:
   "Am I using only what is written below?" If no -- stop and rewrite.

 
ENTITY A (Knowledge Graph 1)
ID: [entity_a_id]
Label: [entity_a_name]
Type(s): [entity_a_types]
Properties: [entity_a_properties]
Relationships (1-2 hops):
[entity_a_relationships]

 
ENTITY B (Knowledge Graph 2)
ID: [entity_b_id]
Label: [entity_b_name]
Type(s): [entity_b_types]
Properties: [entity_b_properties]
Relationships (1-2 hops):
[entity_b_relationships]

 
SYSTEM DECISION: [system_decision]

 
INTERNAL REASONING -- DO NOT INCLUDE IN FINAL OUTPUT
Work through the following steps internally. Be exhaustive and critical.
Final output must contain ONLY the structured block at the end.

 
STEP 1: Structural Profiling
For each entity, derive its structural identity purely from the
evidence above. Do NOT name, recognize, or recall real-world entities.

Entity A:
- Functional role: [What does this entity do or represent based on
  its properties?]
- Distinctive structural signals: [Which attributes/relations are
  rare or specific?]
- Relational fingerprint: [What does its graph neighborhood look like?]
- Ambiguities: [What is underspecified or missing?]

Entity B:
- Functional role: [same]
- Distinctive structural signals: [same]
- Relational fingerprint: [same]
- Ambiguities: [same]

 
STEP 2: Evidence Inventory
List ALL evidence relevant to alignment. Cite exact strings from above.
Do not infer or paraphrase beyond what is written.

SUPPORTING ALIGNMENT:
- [Cite evidence. State why it supports alignment.]
- [Each point must quote or directly reference the evidence text.]

CONTRADICTING ALIGNMENT:
- [Cite evidence. State why it contradicts alignment.]
- [Each point must quote or directly reference the evidence text.]

NEUTRAL / AMBIGUOUS:
- [Evidence that is present but uninformative for this decision.]

 
STEP 3: Evidence Strength Assessment
Classify each piece of evidence by its discriminative power.

STRONG SUPPORT (distinctive, unlikely coincidental):
- [Evidence that would rarely appear in two different entities]

WEAK SUPPORT (generic, coincidental):
- [Evidence that is broad or common]

STRONG CONTRADICTION (irreconcilable conflict):
- [Evidence that makes co-reference structurally impossible]

WEAK CONTRADICTION (explainable by data quality):
- [Minor inconsistency attributable to schema difference or missing
  data]

 
STEP 4: Relational Chain Reasoning
This step is mandatory. Follow any multi-hop chains present in the
evidence.

For each relation chain you can traverse:
- Write out the chain explicitly: A -> rel1 -> X -> rel2 -> Y
- State what conclusion this chain supports or contradicts
- If both entities share a chain endpoint, note it as convergent
  evidence

Example format:
"Entity A: part_of -> [narrative universe X] -> features ->
           [protagonist type Y]
 Entity B: belongs_to -> [saga X] -> centers_on -> [officer type Y]
 -> Both chains converge on [X] and [Y] -> supports alignment"

If no chains are present in the evidence, explicitly state:
"No multi-hop chains available -- decision must rely on direct
attribute comparison only."

 
STEP 5: Decision Analysis
The system decided: [system_decision]

Evaluate this decision against your evidence inventory from Steps 2-4.
- Does the weight of strong evidence support or contradict the
  decision?
- Are there irreconcilable contradictions the system ignored?
- Are there convergent relational chains the system missed?
- Is the decision traceable to the evidence, or does it appear
  arbitrary?
- Would the decision hold if entity labels were replaced with
  anonymous IDs?

That last question is your pre-training bias check. If the answer
is NO -- the system (or judge) relied on label recognition, not
reasoning.

 
STEP 6: Missed Signals
What did the system fail to handle? Choose from:
- Failed to follow relational chains present in evidence
- Over-relied on surface label similarity
- Ignored type/schema conflicts
- Missed convergent graph neighborhood signals
- Over-weighted weak/coincidental evidence
- Under-weighted strong/distinctive evidence
- Failed to notice critical missing evidence on one side
- Reasoned beyond provided evidence (used external knowledge)
- Other: [specify]

Be specific. Reference exact evidence strings.

 
STEP 7: Score Assignment
Score the system's decision from 0-10.

| Score | Meaning                                                    |
|  -|                    |
|  10   | Evidence overwhelmingly supports decision. No missed       |
|       | signals.                                                   |
|   9   | Evidence strongly supports. Minor quibbles only.           |
|   8   | Evidence clearly supports. Slight ambiguity present but    |
|       | irrelevant.                                                |
|   7   | Evidence supports but minor missed nuance. Outcome         |
|       | unaffected.                                                |
|   6   | Evidence moderately supports. Alternate decision also      |
|       | defensible.                                                |
|   5   | Evidence mixed. Decision plausible but weakly supported.   |
|   4   | Evidence weakly supports. Clear warning signs ignored.     |
|   3   | Evidence mostly contradicts. Catchable error.              |
|   2   | Evidence strongly contradicts. Significant obvious error.  |
|   1   | Evidence overwhelmingly contradicts. Near-malfunction.     |
|   0   | Decision defies all evidence. Completely indefensible.     |

Score: ___ / 10

Justification: [Cite specific evidence. Reference rubric tier
explicitly.]

 
STEP 8: Error Category (ONLY if score <= 5)
Assign ONE primary error category:

[ ] Over-relied on surface label similarity
[ ] Ignored type/schema conflict
[ ] Failed to follow relational chains in evidence
[ ] Missed convergent graph neighborhood signals
[ ] Over-weighted weak/coincidental evidence
[ ] Under-weighted strong/distinctive evidence
[ ] Reasoned beyond evidence (external knowledge used)
[ ] Data quality mishandled
[ ] Other: [specify]

 
FINAL OUTPUT -- INCLUDE ONLY THIS BLOCK

ENTITY A: [entity_a_name] ([entity_a_id])
ENTITY B: [entity_b_name] ([entity_b_id])
SYSTEM DECISION: [system_decision]
SCORE: [X]/10
CRITIQUE: [Direct, specific critique of the system decision.
Format: "You [did X with evidence]. You should have [done Y].
This caused [consequence Z]."
Every claim must reference specific evidence strings.
No external knowledge. 3-5 sentences maximum.]
ERROR CATEGORY: [From Step 8, only if score <= 5. Otherwise: N/A]
EVIDENCE USED: [Bullet list of the exact evidence strings that
determined your score. Maximum 5 bullets.]
\end{lstlisting}

\subsection*{A.2 Prompt II: Minimal Evaluation}
Prompt~II is the zero-guidance baseline. It provides no chain-of-thought
instructions, no memory suppression, no evidence boundary, and no rubric.
The judge sees the system decision and scores it using whatever default
reasoning strategy it applies. This represents the condition most
practitioners would use in the absence of prompt engineering, and serves
as the baseline against which Prompt~I measures the total effect of
full prompt design on anchor bias.

\begin{lstlisting}[breaklines=true, breakatwhitespace=true, basicstyle=\ttfamily\footnotesize, frame=single, framesep=5pt]
You are evaluating whether an entity alignment system made a correct
decision.

Two entities from different knowledge graphs are shown below.
The system decided whether they refer to the same real-world entity.
Score how well-justified that decision is given the available
information.

 
ENTITY A (Knowledge Graph 1)
ID: [entity_a_id]
Label: [entity_a_name]
Type(s): [entity_a_types]
Properties: [entity_a_properties]
Relationships: [entity_a_relationships]

 
ENTITY B (Knowledge Graph 2)
ID: [entity_b_id]
Label: [entity_b_name]
Type(s): [entity_b_types]
Properties: [entity_b_properties]
Relationships: [entity_b_relationships]

 
SYSTEM DECISION: [system_decision]

 
Score the system's decision from 0 (completely wrong) to
10 (completely correct).

SCORE: [X]/10
CRITIQUE: [2-3 sentences explaining your score.]
ERROR CATEGORY: N/A
\end{lstlisting}

\subsection*{A.3 Prompt III: LLM as Predictor}
Prompt~III instantiates an \emph{alignment prediction} task rather than
a decision validation task. The system decision is never shown to the
LLM predictor. Instead, the judge forms its own independent alignment prediction
using a 6-step reasoning protocol. Score semantics are inverted relative
to Prompts~I and~II: high scores indicate high confidence in a match,
low scores indicate high confidence in a non-match. The key structural
difference is the physical absence of a label anchor rather than an
instruction to ignore one. This physical absence is what removes the label anchor
structurally, making Prompt~III an alignment-prediction task rather than a debiased
version of decision validation.

\begin{lstlisting}[breaklines=true, breakatwhitespace=true, basicstyle=\ttfamily\footnotesize, frame=single, framesep=5pt]
SYSTEM ROLE
You are an Entity Alignment Predictor. Your task is to assess whether
two entities from different knowledge graphs refer to the same
real-world entity. You have no system decision to evaluate -- you must
form your own judgment from the evidence provided.

HARD CONSTRAINTS:
1. EVIDENCE BOUNDARY: Reason EXCLUSIVELY from the evidence provided
   below.
2. MEMORY SUPPRESSION: Do NOT use external world knowledge or prior
   training knowledge about these entities. If you recognise an entity,
   DISCARD that recognition. Treat every entity as previously unknown.
3. SCORE SEMANTICS: Higher score = higher confidence they are the SAME
   entity.
   10 = near-certain match | 5 = ambiguous | 0 = near-certain non-match.

 
ENTITY A (Knowledge Graph 1)
ID: [entity_a_id]
Label: [entity_a_name]
Type(s): [entity_a_types]
Properties: [entity_a_properties]
Relationships (1-2 hops):
[entity_a_relationships]

 
ENTITY B (Knowledge Graph 2)
ID: [entity_b_id]
Label: [entity_b_name]
Type(s): [entity_b_types]
Properties: [entity_b_properties]
Relationships (1-2 hops):
[entity_b_relationships]

 
INTERNAL REASONING -- DO NOT INCLUDE IN FINAL OUTPUT

 
STEP 1: Structural Profiling
Derive the structural identity of each entity purely from the evidence
above. Do NOT name, recognise, or recall real-world entities.

Entity A:
- Functional role: [what does it represent based on its properties?]
- Distinctive signals: [which attributes/relations are rare or specific?]
- Relational fingerprint: [what does its graph neighbourhood look like?]
- Ambiguities: [what is underspecified or missing?]

Entity B:
- Functional role: [same]
- Distinctive signals: [same]
- Relational fingerprint: [same]
- Ambiguities: [same]

 
STEP 2: Evidence Alignment Inventory
Cite exact strings from above. Do not infer beyond what is written.

SUPPORTING MATCH:
- [Quote evidence. State why it supports co-reference.]

CONTRADICTING MATCH:
- [Quote evidence. State why it contradicts co-reference.]

NEUTRAL / AMBIGUOUS:
- [Evidence present but uninformative for this decision.]

 
STEP 3: Relational Chain Reasoning
Follow any multi-hop chains present in the evidence.

For each chain: A -> rel1 -> X -> rel2 -> Y
State what conclusion this chain supports or contradicts.

If no chains present: "No multi-hop chains available."

 
STEP 4: Evidence Strength Assessment
STRONG MATCH SIGNALS (distinctive, unlikely coincidental):
- [Evidence that would rarely appear in two different entities]

WEAK MATCH SIGNALS (generic, coincidental):
- [Evidence that is broad or common across many entities]

STRONG NON-MATCH SIGNALS (irreconcilable conflict):
- [Evidence that makes co-reference structurally impossible]

WEAK NON-MATCH SIGNALS (explainable by schema difference):
- [Minor inconsistency attributable to data quality or schema mismatch]

 
STEP 5: Alignment Score Assignment
Score 0-10 reflecting your confidence that A and B refer to the same
entity.

| Score | Meaning                                                       |
|  -|                --|
|  10   | Near-certain match. Evidence overwhelmingly supports          |
|       | co-reference.                                                 |
|   9   | Strong match. Evidence strongly supports. Minor gaps only.    |
|   8   | Likely match. Clear convergence with slight ambiguity.        |
|   7   | Probable match. More supporting than contradicting evidence.  |
|   6   | Lean match. Weak convergence, alternate explanation possible. |
|   5   | Ambiguous. Evidence insufficient or perfectly balanced.       |
|   4   | Lean non-match. More contradicting than supporting.           |
|   3   | Probable non-match. Structural conflicts present.             |
|   2   | Likely non-match. Evidence strongly diverges.                 |
|   1   | Strong non-match. Near-irreconcilable conflict.               |
|   0   | Near-certain non-match. Evidence overwhelmingly contradicts   |
|       | co-reference.                                                 |

Bias check: Would your score change if entity labels were replaced with
anonymous IDs? If YES -- you are using name recognition, not evidence.
Adjust your score to reflect evidence only.

Score: ___ / 10

Justification: [Cite specific evidence strings. 3-5 sentences maximum.]

 
STEP 6: Error Category (ONLY if score is 4 or below)
[ ] Type/schema conflict
[ ] Contradicting relational chains
[ ] Irreconcilable attribute conflict
[ ] Insufficient evidence on one or both sides
[ ] Other: [specify]

 
FINAL OUTPUT -- INCLUDE ONLY THIS BLOCK

ENTITY A: [entity_a_name] ([entity_a_id])
ENTITY B: [entity_b_name] ([entity_b_id])
PREDICTION: [MATCH / NON-MATCH / UNCERTAIN]
SCORE: [X]/10
CRITIQUE: [Direct justification of your alignment prediction.
Format: "The evidence [supports/contradicts/is insufficient for]
alignment because [specific evidence strings]. The strongest signal
was [X]."
3-5 sentences. No external knowledge. Cite evidence strings only.]
ERROR CATEGORY: [From Step 6 if score <= 4. Otherwise: N/A]
EVIDENCE USED: [Bullet list of exact evidence strings that most
influenced your score. Maximum 5 bullets.]
\end{lstlisting}

\paragraph{Structural distinction across prompts.}
Prompts~I and~II are \emph{decision validation} tasks: the LLM
acts as a judge, receiving a system label (\textsc{Match} or
\textsc{Non-Match}) and scoring how justified that decision is.
Prompt~III is an \emph{alignment prediction} task: the LLM acts
as a predictor; no label is shown and its score directly encodes
its own alignment confidence (high~$=$~match, low~$=$~non-match). The mechanistic
basis for Prompt~III's revelation of underlying capability is a
structural difference: the physical absence of the label anchor rather than an
instruction to ignore it. This mechanism is empirically demonstrated in
the results above, particularly in the perturbation bias analysis.

\section*{Appendix B: Judge Meta-Evaluation Threshold Sweeps}
\label{app:thresholds}

Tables~\ref{tab:thresh_easy_dw}, \ref{tab:thresh_easy_emb_dw},
\ref{tab:thresh_easy_dy},
and~\ref{tab:thresh_easy_bio} report J-Precision, J-Recall,
J-F1, and J-ROC-AUC at thresholds $t \in \{5,6,7,8,9\}$
for EasyEA \cite{cheng-etal-2025-easyea} across DW, DY, and BIO respectively.
Formally, $c_i = \mathbb{1}[\hat{y}_i = y_i]$ is the correctness
indicator computed after sampling; J-Precision, J-Recall, J-F1,
and J-ROC-AUC are computed between $\mathbf{c}$ and the judge's
binary acceptance $y_{\text{judge}}^{(i)}(t) =
\mathbb{1}[s^{(i)} \geq t]$, measuring whether the judge correctly
identifies EA system correctness.
Table~\ref{tab:thresh_neu} reports the same for NeuSymEA \cite{chen2025neusymea}
and NeuSymEA-NoRefine.

\begin{table}[!htbp]
\centering
\tiny
\setlength{\tabcolsep}{1.5pt}
\renewcommand{\arraystretch}{0.68}
\caption{EasyEA \cite{cheng-etal-2025-easyea} threshold sweep (DW): J-Prec, J-Rec, J-F1 at $t\in\{5,\ldots,9\}$, J-ROC-AUC, $n{=}500$.}
\label{tab:thresh_easy_dw}
\resizebox{\columnwidth}{!}{%
\begin{tabular}{llrrrrc}
\toprule
\textbf{Judge} & \textbf{Prompt} & \textbf{$t$} &
\textbf{J-Prec} & \textbf{J-Rec} & \textbf{J-F1} & \textbf{J-ROC-AUC} \\
\midrule
\multirow{5}{*}{GPT-4o-mini}
  & \multirow{5}{*}{I}
    & 5 & 0.506 & 0.980 & 0.668 & \multirow{5}{*}{0.528} \\
& & 6 & 0.500 & 0.828 & 0.623 & \\
& & 7 & 0.505 & 0.584 & 0.542 & \\
& & 8 & 0.528 & 0.536 & 0.532 & \\
& & 9 & 0.579 & 0.264 & 0.363 & \\
\cmidrule(l){2-7}
& \multirow{5}{*}{II}
    & 5 & 0.491 & 0.964 & 0.650 & \multirow{5}{*}{0.244} \\
& & 6 & 0.490 & 0.960 & 0.649 & \\
& & 7 & 0.489 & 0.956 & 0.647 & \\
& & 8 & 0.490 & 0.956 & 0.648 & \\
& & 9 & 0.474 & 0.860 & 0.611 & \\
\cmidrule(l){2-7}
& \multirow{5}{*}{III}
    & 5 & 0.968 & 0.976 & 0.972 & \multirow{5}{*}{---} \\
& & 6 & 0.972 & 0.976 & 0.974 & \\
& & 7 & 0.972 & 0.976 & 0.974 & \\
& & 8 & 0.972 & 0.968 & 0.970 & \\
& & 9 & 1.000 & 0.856 & 0.922 & \\
\midrule
\multirow{5}{*}{GPT-4o}
  & \multirow{5}{*}{I}
    & 5 & 0.490 & 0.971 & 0.652 & \multirow{5}{*}{0.464} \\
& & 6 & 0.468 & 0.854 & 0.604 & \\
& & 7 & 0.463 & 0.753 & 0.573 & \\
& & 8 & 0.479 & 0.711 & 0.572 & \\
& & 9 & 0.514 & 0.536 & 0.525 & \\
\cmidrule(l){2-7}
& \multirow{5}{*}{II}
    & 5 & 0.497 & 0.956 & 0.654 & \multirow{5}{*}{0.226} \\
& & 6 & 0.496 & 0.948 & 0.651 & \\
& & 7 & 0.489 & 0.920 & 0.639 & \\
& & 8 & 0.472 & 0.848 & 0.607 & \\
& & 9 & 0.431 & 0.688 & 0.530 & \\
\cmidrule(l){2-7}
& \multirow{5}{*}{III}
    & 5 & 0.968 & 0.972 & 0.970 & \multirow{5}{*}{---} \\
& & 6 & 0.992 & 0.948 & 0.969 & \\
& & 7 & 0.992 & 0.948 & 0.969 & \\
& & 8 & 0.991 & 0.924 & 0.957 & \\
& & 9 & 0.995 & 0.780 & 0.874 & \\
\midrule
\multirow{5}{*}{Claude Opus 4.7}
  & \multirow{5}{*}{I}
    & 5 & 0.492 & 0.956 & 0.649 & \multirow{5}{*}{0.342} \\
& & 6 & 0.479 & 0.908 & 0.627 & \\
& & 7 & 0.469 & 0.844 & 0.603 & \\
& & 8 & 0.460 & 0.812 & 0.588 & \\
& & 9 & 0.399 & 0.496 & 0.442 & \\
\cmidrule(l){2-7}
& \multirow{5}{*}{II}
    & 5 & 0.497 & 0.968 & 0.657 & \multirow{5}{*}{0.494} \\
& & 6 & 0.497 & 0.968 & 0.657 & \\
& & 7 & 0.497 & 0.964 & 0.656 & \\
& & 8 & 0.497 & 0.964 & 0.656 & \\
& & 9 & 0.499 & 0.936 & 0.651 & \\
\cmidrule(l){2-7}
& \multirow{5}{*}{III}
    & 5 & 0.946 & 0.988 & 0.967 & \multirow{5}{*}{---} \\
& & 6 & 0.968 & 0.972 & 0.970 & \\
& & 7 & 0.984 & 0.960 & 0.972 & \\
& & 8 & 0.983 & 0.948 & 0.965 & \\
& & 9 & 1.000 & 0.859 & 0.924 & \\
\bottomrule
\end{tabular}%
}
\end{table}

\begin{table}[!htbp]
\centering
\tiny
\setlength{\tabcolsep}{1.5pt}
\renewcommand{\arraystretch}{0.62}
\caption{EasyEA-EmbOnly threshold sweep (DW): J-Prec, J-Rec, J-F1 at $t\in\{5,\ldots,9\}$, J-ROC-AUC, $n{=}500$.}
\label{tab:thresh_easy_emb_dw}
\resizebox{\columnwidth}{!}{%
\begin{tabular}{llrrrrc}
\toprule
\textbf{Judge} & \textbf{Prompt} & \textbf{$t$} &
\textbf{J-Prec} & \textbf{J-Rec} & \textbf{J-F1} & \textbf{J-ROC-AUC} \\
\midrule
\multirow{5}{*}{GPT-4o-mini}
  & \multirow{5}{*}{I}
    & 5 & 0.503 & 0.968 & 0.662 & \multirow{5}{*}{0.533} \\
& & 6 & 0.480 & 0.760 & 0.588 & \\
& & 7 & 0.527 & 0.592 & 0.557 & \\
& & 8 & 0.554 & 0.536 & 0.545 & \\
& & 9 & 0.589 & 0.264 & 0.365 & \\
\cmidrule(l){2-7}
& \multirow{5}{*}{II}
    & 5 & 0.490 & 0.960 & 0.649 & \multirow{5}{*}{0.295} \\
& & 6 & 0.487 & 0.948 & 0.643 & \\
& & 7 & 0.481 & 0.928 & 0.634 & \\
& & 8 & 0.480 & 0.924 & 0.632 & \\
& & 9 & 0.469 & 0.840 & 0.602 & \\
\cmidrule(l){2-7}
& \multirow{5}{*}{III}
    & 5 & 0.941 & 0.964 & 0.953 & \multirow{5}{*}{---} \\
& & 6 & 0.945 & 0.964 & 0.954 & \\
& & 7 & 0.945 & 0.956 & 0.950 & \\
& & 8 & 0.948 & 0.956 & 0.952 & \\
& & 9 & 0.990 & 0.828 & 0.902 & \\
\midrule
\multirow{5}{*}{GPT-4o}
  & \multirow{5}{*}{I}
    & 5 & 0.487 & 0.966 & 0.648 & \multirow{5}{*}{0.445} \\
& & 6 & 0.467 & 0.860 & 0.605 & \\
& & 7 & 0.470 & 0.797 & 0.591 & \\
& & 8 & 0.465 & 0.703 & 0.560 & \\
& & 9 & 0.481 & 0.479 & 0.480 & \\
\cmidrule(l){2-7}
& \multirow{5}{*}{II}
    & 5 & 0.502 & 0.964 & 0.660 & \multirow{5}{*}{0.235} \\
& & 6 & 0.497 & 0.944 & 0.651 & \\
& & 7 & 0.491 & 0.920 & 0.641 & \\
& & 8 & 0.481 & 0.872 & 0.620 & \\
& & 9 & 0.453 & 0.728 & 0.558 & \\
\cmidrule(l){2-7}
& \multirow{5}{*}{III}
    & 5 & 0.996 & 0.964 & 0.980 & \multirow{5}{*}{---} \\
& & 6 & 0.996 & 0.940 & 0.967 & \\
& & 7 & 0.996 & 0.940 & 0.967 & \\
& & 8 & 0.996 & 0.908 & 0.950 & \\
& & 9 & 1.000 & 0.788 & 0.881 & \\
\midrule
\multirow{5}{*}{Claude Opus 4.7}
  & \multirow{5}{*}{I}
    & 5 & 0.501 & 0.964 & 0.659 & \multirow{5}{*}{0.292} \\
& & 6 & 0.488 & 0.916 & 0.637 & \\
& & 7 & 0.459 & 0.812 & 0.587 & \\
& & 8 & 0.457 & 0.804 & 0.583 & \\
& & 9 & 0.385 & 0.496 & 0.434 & \\
\cmidrule(l){2-7}
& \multirow{5}{*}{II}
    & 5 & 0.488 & 0.948 & 0.644 & \multirow{5}{*}{0.439} \\
& & 6 & 0.488 & 0.948 & 0.644 & \\
& & 7 & 0.488 & 0.948 & 0.644 & \\
& & 8 & 0.488 & 0.948 & 0.644 & \\
& & 9 & 0.489 & 0.936 & 0.642 & \\
\cmidrule(l){2-7}
& \multirow{5}{*}{III}
    & 5 & 1.000 & 0.984 & 0.992 & \multirow{5}{*}{---} \\
& & 6 & 1.000 & 0.960 & 0.980 & \\
& & 7 & 1.000 & 0.960 & 0.980 & \\
& & 8 & 1.000 & 0.936 & 0.967 & \\
& & 9 & 1.000 & 0.820 & 0.901 & \\
\bottomrule
\end{tabular}}
\end{table}

\begin{table}[!htbp]
\centering
\tiny
\setlength{\tabcolsep}{1.5pt}
\renewcommand{\arraystretch}{0.68}
\caption{EasyEA \cite{cheng-etal-2025-easyea} threshold sweep (DY): J-Prec, J-Rec, J-F1 at $t\in\{5,\ldots,9\}$, J-ROC-AUC, $n{=}500$.}
\label{tab:thresh_easy_dy}
\resizebox{\columnwidth}{!}{%
\begin{tabular}{llrrrrc}
\toprule
\textbf{Judge} & \textbf{Prompt} & \textbf{$t$} &
\textbf{J-Prec} & \textbf{J-Rec} & \textbf{J-F1} & \textbf{J-ROC-AUC} \\
\midrule
\multirow{5}{*}{GPT-4o-mini}
  & \multirow{5}{*}{I}
    & 5 & 0.588 & 0.992 & 0.738 & \multirow{5}{*}{0.567} \\
& & 6 & 0.565 & 0.840 & 0.675 & \\
& & 7 & 0.554 & 0.720 & 0.626 & \\
& & 8 & 0.549 & 0.668 & 0.603 & \\
& & 9 & 0.509 & 0.460 & 0.483 & \\
\cmidrule(l){2-7}
& \multirow{5}{*}{II}
    & 5 & 0.678 & 0.992 & 0.805 & \multirow{5}{*}{0.712} \\
& & 6 & 0.678 & 0.992 & 0.805 & \\
& & 7 & 0.676 & 0.984 & 0.801 & \\
& & 8 & 0.669 & 0.952 & 0.785 & \\
& & 9 & 0.673 & 0.840 & 0.747 & \\
\cmidrule(l){2-7}
& \multirow{5}{*}{III}
    & 5 & 0.576 & 1.000 & 0.731 & \multirow{5}{*}{0.527} \\
& & 6 & 0.585 & 1.000 & 0.739 & \\
& & 7 & 0.585 & 1.000 & 0.739 & \\
& & 8 & 0.589 & 0.996 & 0.740 & \\
& & 9 & 0.598 & 0.940 & 0.731 & \\
\midrule
\multirow{5}{*}{GPT-4o}
  & \multirow{5}{*}{I}
    & 5 & 0.701 & 1.000 & 0.824 & \multirow{5}{*}{0.783} \\
& & 6 & 0.728 & 0.910 & 0.809 & \\
& & 7 & 0.731 & 0.725 & 0.728 & \\
& & 8 & 0.719 & 0.586 & 0.646 & \\
& & 9 & 0.637 & 0.295 & 0.403 & \\
\cmidrule(l){2-7}
& \multirow{5}{*}{II}
    & 5 & 0.714 & 1.000 & 0.833 & \multirow{5}{*}{0.750} \\
& & 6 & 0.719 & 0.992 & 0.834 & \\
& & 7 & 0.718 & 0.968 & 0.825 & \\
& & 8 & 0.717 & 0.892 & 0.795 & \\
& & 9 & 0.714 & 0.660 & 0.686 & \\
\cmidrule(l){2-7}
& \multirow{5}{*}{III}
    & 5 & 0.577 & 1.000 & 0.732 & \multirow{5}{*}{0.468} \\
& & 6 & 0.611 & 0.992 & 0.756 & \\
& & 7 & 0.611 & 0.992 & 0.756 & \\
& & 8 & 0.611 & 0.960 & 0.747 & \\
& & 9 & 0.573 & 0.752 & 0.651 & \\
\midrule
\multirow{5}{*}{Claude Opus 4.7}
  & \multirow{5}{*}{I}
    & 5 & 0.773 & 0.992 & 0.869 & \multirow{5}{*}{0.766} \\
& & 6 & 0.776 & 0.968 & 0.861 & \\
& & 7 & 0.737 & 0.728 & 0.732 & \\
& & 8 & 0.733 & 0.712 & 0.722 & \\
& & 9 & 0.465 & 0.188 & 0.268 & \\
\cmidrule(l){2-7}
& \multirow{5}{*}{II}
    & 5 & 0.778 & 0.996 & 0.874 & \multirow{5}{*}{0.866} \\
& & 6 & 0.778 & 0.996 & 0.874 & \\
& & 7 & 0.778 & 0.996 & 0.874 & \\
& & 8 & 0.777 & 0.992 & 0.872 & \\
& & 9 & 0.786 & 0.984 & 0.874 & \\
\cmidrule(l){2-7}
& \multirow{5}{*}{III}
    & 5 & 0.572 & 1.000 & 0.728 & \multirow{5}{*}{0.507} \\
& & 6 & 0.572 & 1.000 & 0.728 & \\
& & 7 & 0.586 & 0.984 & 0.734 & \\
& & 8 & 0.581 & 0.960 & 0.724 & \\
& & 9 & 0.555 & 0.732 & 0.631 & \\
\bottomrule
\end{tabular}}
\end{table}

\begin{table}[!htbp]
\centering
\tiny
\setlength{\tabcolsep}{1.5pt}
\renewcommand{\arraystretch}{0.68}
\caption{EasyEA \cite{cheng-etal-2025-easyea} threshold sweep (BIO): J-Prec, J-Rec, J-F1 at $t\in\{5,\ldots,9\}$, J-ROC-AUC, $n{=}500$.}
\label{tab:thresh_easy_bio}
\resizebox{\columnwidth}{!}{%
\begin{tabular}{llrrrrc}
\toprule
\textbf{Judge} & \textbf{Prompt} & \textbf{$t$} &
\textbf{J-Prec} & \textbf{J-Rec} & \textbf{J-F1} & \textbf{J-ROC-AUC} \\
\midrule
\multirow{5}{*}{GPT-4o-mini}
  & \multirow{5}{*}{I}
    & 5 & 0.499 & 0.948 & 0.654 & \multirow{5}{*}{0.477} \\
& & 6 & 0.474 & 0.756 & 0.582 & \\
& & 7 & 0.474 & 0.520 & 0.496 & \\
& & 8 & 0.473 & 0.456 & 0.464 & \\
& & 9 & 0.539 & 0.304 & 0.389 & \\
\cmidrule(l){2-7}
& \multirow{5}{*}{II}
    & 5 & 0.493 & 0.928 & 0.644 & \multirow{5}{*}{0.341} \\
& & 6 & 0.488 & 0.912 & 0.636 & \\
& & 7 & 0.470 & 0.848 & 0.605 & \\
& & 8 & 0.453 & 0.772 & 0.571 & \\
& & 9 & 0.436 & 0.600 & 0.505 & \\
\cmidrule(l){2-7}
& \multirow{5}{*}{III}
    & 5 & 0.865 & 0.900 & 0.882 & \multirow{5}{*}{0.931} \\
& & 6 & 0.885 & 0.860 & 0.872 & \\
& & 7 & 0.885 & 0.860 & 0.872 & \\
& & 8 & 0.887 & 0.820 & 0.852 & \\
& & 9 & 0.938 & 0.548 & 0.692 & \\
\midrule
\multirow{5}{*}{GPT-4o}
  & \multirow{5}{*}{I}
    & 5 & 0.493 & 0.966 & 0.653 & \multirow{5}{*}{0.423} \\
& & 6 & 0.478 & 0.803 & 0.600 & \\
& & 7 & 0.468 & 0.624 & 0.535 & \\
& & 8 & 0.449 & 0.491 & 0.469 & \\
& & 9 & 0.358 & 0.184 & 0.243 & \\
\cmidrule(l){2-7}
& \multirow{5}{*}{II}
    & 5 & 0.501 & 0.928 & 0.651 & \multirow{5}{*}{0.295} \\
& & 6 & 0.498 & 0.912 & 0.644 & \\
& & 7 & 0.492 & 0.884 & 0.632 & \\
& & 8 & 0.474 & 0.788 & 0.592 & \\
& & 9 & 0.421 & 0.540 & 0.473 & \\
\cmidrule(l){2-7}
& \multirow{5}{*}{III}
    & 5 & 0.865 & 0.952 & 0.907 & \multirow{5}{*}{0.945} \\
& & 6 & 0.887 & 0.908 & 0.897 & \\
& & 7 & 0.890 & 0.908 & 0.899 & \\
& & 8 & 0.898 & 0.848 & 0.872 & \\
& & 9 & 0.930 & 0.588 & 0.721 & \\
\midrule
\multirow{5}{*}{Claude Opus 4.7}
  & \multirow{5}{*}{I}
    & 5 & 0.503 & 0.924 & 0.652 & \multirow{5}{*}{0.223} \\
& & 6 & 0.489 & 0.872 & 0.626 & \\
& & 7 & 0.447 & 0.696 & 0.545 & \\
& & 8 & 0.427 & 0.636 & 0.511 & \\
& & 9 & 0.045 & 0.028 & 0.034 & \\
\cmidrule(l){2-7}
& \multirow{5}{*}{II}
    & 5 & 0.514 & 0.964 & 0.670 & \multirow{5}{*}{0.457} \\
& & 6 & 0.514 & 0.964 & 0.670 & \\
& & 7 & 0.503 & 0.919 & 0.650 & \\
& & 8 & 0.501 & 0.891 & 0.641 & \\
& & 9 & 0.501 & 0.802 & 0.617 & \\
\cmidrule(l){2-7}
& \multirow{5}{*}{III}
    & 5 & 0.859 & 0.948 & 0.901 & \multirow{5}{*}{0.948} \\
& & 6 & 0.883 & 0.936 & 0.909 & \\
& & 7 & 0.904 & 0.868 & 0.886 & \\
& & 8 & 0.907 & 0.860 & 0.883 & \\
& & 9 & 0.929 & 0.680 & 0.785 & \\
\bottomrule
\end{tabular}}
\end{table}

\begin{table*}[!htbp]
\centering
\footnotesize
\setlength{\tabcolsep}{1.2pt}
\renewcommand{\arraystretch}{0.75}
\caption{NeuSymEA \cite{chen2025neusymea} and NeuSymEA-NoRefine best-threshold summary:
J-Prec, J-Rec, J-F1, J-ROC-AUC, and Score Gap at
$t^*=\arg\max_t\text{J-F1}(t)$ across DW and DY
($n{=}500$; BIO excluded, NeuSymEA \cite{chen2025neusymea} not evaluated on BIO).
Discriminates: Yes = J-ROC-AUC~$>0.6$;
Invert = J-ROC-AUC~$<0.45$; $\sim$Chance otherwise.}
\label{tab:thresh_neu}
\begin{tabular}{llllrrrrrr}
\toprule
\textbf{Dataset} & \textbf{System} & \textbf{Judge} &
\textbf{Prompt} & \textbf{$t^*$} & \textbf{J-Prec} &
\textbf{J-Rec} & \textbf{J-F1} & \textbf{J-ROC-AUC} &
\textbf{Score Gap} \\
\midrule
\multirow{18}{*}{DW}
& \multirow{9}{*}{NeuSymEA}
  & GPT-4o-mini & I  & 5 & 0.508 & 0.972 & 0.668 & 0.355 & $-$0.900 \\
& & GPT-4o-mini & II & 5 & 0.494 & 0.972 & 0.655 & 0.145 & $-$1.072 \\
& & GPT-4o-mini & III& 6 & 0.972 & 0.988 & 0.980 & 0.981 & $+$8.248 \\
& & GPT-4o      & I  & 5 & 0.496 & 0.983 & 0.659 & 0.251 & $-$1.207 \\
& & GPT-4o      & II & 5 & 0.503 & 0.992 & 0.668 & 0.165 & $-$1.000 \\
& & GPT-4o      & III& 5 & 0.976 & 0.988 & 0.982 & 0.995 & $+$8.300 \\
& & Claude Opus 4.7 & I  & 5 & 0.502 & 0.984 & 0.665 & 0.189 & $-$1.252 \\
& & Claude Opus 4.7 & II & 7 & 0.502 & 0.984 & 0.665 & 0.451 & $-$0.115 \\
& & Claude Opus 4.7 & III& 7 & 0.992 & 0.976 & 0.984 & 0.995 & $+$8.320 \\
\cmidrule(l){2-10}
& \multirow{9}{*}{NeuSymEA-NoRefine}
  & GPT-4o-mini & I  & 5 & 0.494 & 0.970 & 0.654 & 0.380 & $-$0.725 \\
& & GPT-4o-mini & II & 5 & 0.499 & 0.988 & 0.663 & 0.142 & $-$0.948 \\
& & GPT-4o-mini & III& 6 & 0.969 & 0.988 & 0.978 & 0.987 & $+$8.248 \\
& & GPT-4o      & I  & 5 & 0.490 & 0.983 & 0.654 & 0.220 & $-$1.357 \\
& & GPT-4o      & II & 5 & 0.497 & 0.976 & 0.659 & 0.137 & $-$1.132 \\
& & GPT-4o      & III& 5 & 0.980 & 0.984 & 0.982 & 0.996 & $+$8.284 \\
& & Claude Opus 4.7 & I  & 5 & 0.502 & 0.988 & 0.666 & 0.168 & $-$1.292 \\
& & Claude Opus 4.7 & II & 5 & 0.502 & 0.992 & 0.667 & 0.447 & $-$0.080 \\
& & Claude Opus 4.7 & III& 5 & 0.976 & 0.996 & 0.986 & 0.996 & $+$8.424 \\
\midrule
\multirow{18}{*}{DY}
& \multirow{9}{*}{NeuSymEA}
  & GPT-4o-mini & I  & 5 & 0.502 & 0.984 & 0.665 & 0.509 & $+$0.028 \\
& & GPT-4o-mini & II & 5 & 0.500 & 1.000 & 0.667 & 0.197 & $-$0.748 \\
& & GPT-4o-mini & III& 8 & 1.000 & 1.000 & 1.000 & 1.000 & $+$9.132 \\
& & GPT-4o      & I  & 5 & 0.502 & 0.996 & 0.668 & 0.243 & $-$1.183 \\
& & GPT-4o      & II & 6 & 0.501 & 1.000 & 0.668 & 0.123 & $-$1.092 \\
& & GPT-4o      & III& 6 & 0.996 & 1.000 & 0.998 & 0.999 & $+$8.452 \\
& & Claude Opus 4.7 & I  & 5 & 0.499 & 1.000 & 0.666 & 0.129 & $-$1.602 \\
& & Claude Opus 4.7 & II & 5 & 0.501 & 1.000 & 0.668 & 0.477 & $-$0.016 \\
& & Claude Opus 4.7 & III& 6 & 0.996 & 0.996 & 0.996 & 1.000 & $+$7.932 \\
\cmidrule(l){2-10}
& \multirow{9}{*}{NeuSymEA-NoRefine}
  & GPT-4o-mini & I  & 5 & 0.500 & 1.000 & 0.667 & 0.468 & $-$0.260 \\
& & GPT-4o-mini & II & 5 & 0.498 & 0.992 & 0.663 & 0.212 & $-$0.724 \\
& & GPT-4o-mini & III& 5 & 0.996 & 1.000 & 0.998 & 1.000 & $+$9.135 \\
& & GPT-4o      & I  & 5 & 0.499 & 0.992 & 0.664 & 0.244 & $-$1.184 \\
& & GPT-4o      & II & 5 & 0.501 & 1.000 & 0.668 & 0.141 & $-$0.967 \\
& & GPT-4o      & III& 6 & 0.996 & 1.000 & 0.998 & 0.999 & $+$8.412 \\
& & Claude Opus 4.7 & I  & 5 & 0.501 & 1.000 & 0.668 & 0.132 & $-$1.605 \\
& & Claude Opus 4.7 & II & 5 & 0.500 & 1.000 & 0.667 & 0.458 & $-$0.088 \\
& & Claude Opus 4.7 & III& 5 & 0.992 & 1.000 & 0.996 & 1.000 & $+$7.944 \\
\bottomrule
\end{tabular}
\end{table*}

\section*{Appendix C: Computational Cost and API Usage}
\label{app:cost}

Table~\ref{tab:cost} reports the complete API usage and cost breakdown across Experiments~1 and~2. Experiment~1 comprises 108,000 calls covering 3 datasets $\times$ 3 prompts $\times$ 3 judges $\times$ 4,000 pairs per run (original $+$ 7 perturbation strategies $\times$ 500 pairs). Experiment~2 comprises 45,000 calls: 36,000 for DW and DY (2 datasets $\times$ (2 EA systems $+$ 2 ablations) $\times$ 3 prompts $\times$ 3 judges $\times$ 500 pairs) and 9,000 for BIO (1 EA system with 1 ablation only, since NeuSymEA \cite{chen2025neusymea} was not evaluated on BIO, $\times$ 3 prompts $\times$ 3 judges $\times$ 500 pairs). Token counts for GPT-4o-mini and GPT-4o in Experiment~1 are exact; all remaining counts are derived by 50-row sampling extrapolated to the full condition size. API calls were made via OpenRouter with temperature 0 and
\texttt{max\_completion\_tokens}=1500 across all conditions.
All calls within each experiment were completed within a
single continuous period, minimizing the risk of model
snapshot changes; moreover, temperature~$=0$ ensures fully
deterministic outputs for any fixed model version, though we
note in \S\ref{sec:experiments} that this does not by itself
formally guarantee determinism across arbitrary snapshots.
OpenRouter response metadata confirms direct routing to each
provider's official endpoint throughout (see \S\ref{sec:experiments}).
Pricing as of May 2026: GPT-4o-mini \$0.15/\$0.60, GPT-4o \$2.50/\$10.00, Claude Opus 4.7 \$5.00/\$25.00 per million input/output tokens. Total expenditure across Experiments~1 and~2 was \textbf{\$577.17} over \textbf{153,000 calls} and \textbf{238.8M tokens}. The NeuSymEA \cite{chen2025neusymea} label-flip extension (Experiment~3 replication, 2,400 additional calls on DW and DY) added approximately \$8--12 at the same per-call rates. The human evaluation (Experiment~4) required no additional API calls, as it reused judge scores already collected in Experiments~2--3 alongside newly collected human annotations.

\begin{table}[!htbp]
\centering
\scriptsize
\setlength{\tabcolsep}{0.8pt}
\renewcommand{\arraystretch}{0.7}
\caption{API usage and cost by experiment, dataset, prompt, model.}
\label{tab:cost}
\resizebox{\columnwidth}{!}{%
\begin{tabular}{llrrrrr}
\toprule
\textbf{Exp.} & \textbf{Cond.} & \textbf{Calls} & \textbf{In(M)} & \textbf{Out(M)} & \textbf{Cost} & \textbf{Avg} \\
\midrule
\multirow{9}{*}{Exp.~1}
& DW-I & 12,000 & 21.4 &  9.8 & \$68.47  & \$0.0057 \\
& DW-II & 12,000 &  4.4 &  4.2 & \$21.92  & \$0.0018 \\
& DW-III & 12,000 & 17.0 &  9.7 & \$57.09  & \$0.0048 \\
& DY-I & 12,000 & 21.8 & 10.2 & \$70.70  & \$0.0059 \\
& DY-II & 12,000 &  4.3 &  3.9 & \$20.65  & \$0.0017 \\
& DY-III & 12,000 & 16.3 &  9.4 & \$57.53  & \$0.0048 \\
& BIO-I & 12,000 & 21.9 & 10.4 & \$69.22  & \$0.0058 \\
& BIO-II & 12,000 &  4.4 &  4.2 & \$22.87  & \$0.0019 \\
& BIO-III & 12,000 & 16.5 &  9.8 & \$52.79  & \$0.0044 \\
\cmidrule(l){2-7}
& \textbf{Sub} & \textbf{108k} & \textbf{128.0} & \textbf{71.6} & \textbf{\$441.24} & \textbf{\$0.004} \\
\midrule
\multirow{9}{*}{Exp.~2}
& DW-I & 6,000 & 3.7 & 2.3 & \$20.18 & \$0.0034 \\
& DW-II & 6,000 & 3.7 & 2.0 & \$15.54 & \$0.0026 \\
& DW-III & 6,000 & 3.7 & 2.3 & \$19.39 & \$0.0032 \\
& DY-I & 6,000 & 3.4 & 2.3 & \$19.59 & \$0.0033 \\
& DY-II & 6,000 & 3.4 & 2.0 & \$14.86 & \$0.0025 \\
& DY-III & 6,000 & 3.7 & 2.3 & \$18.54 & \$0.0031 \\
& BIO-I & 3,000 & 0.9 & 0.6 & \$10.27 & \$0.0034 \\
& BIO-II & 3,000 & 0.9 & 0.5 & \$7.86 & \$0.0026 \\
& BIO-III & 3,000 & 0.9 & 0.6 & \$9.70 & \$0.0032 \\
\cmidrule(l){2-7}
& \textbf{Sub} & \textbf{45k} & \textbf{24.3} & \textbf{14.9} & \textbf{\$135.93} & \textbf{\$0.003} \\
\midrule
\textbf{Total} & & \textbf{153k} & \textbf{152.3} & \textbf{86.5} & \textbf{\$577.17} & \\
\bottomrule
\end{tabular}}
\end{table}

\section*{Appendix D: Robustness and Statistical Validation}
\label{app:robustness}

\subsection*{D.0 Sample Size Justification}

The choice of $n=500$ pairs per condition (rather than all 15K available pairs) is justified by three factors. First, \emph{statistical convergence}: 95\% bootstrap confidence intervals for PBI and MRI (Appendix D.2) show tight bands at $n=500$ that separate BIO from the general-domain datasets in all but one condition, indicating stable point estimates and sufficient precision to detect the observed domain asymmetry. Second, \emph{computational feasibility}: $n=500$ requires approximately 4,000 API calls per Experiment 1 run (8 perturbations per pair) across 3 judges and 3 datasets; using all 15K pairs would exceed available compute and API resources. Third, \emph{metric robustness}: PBI, JSR, Score Gap, and J-ROC-AUC are either distribution-free (rank-based) or converge quickly to stable values with sample size. The directional findings (label-exposing prompts fail, label-free Prompt~III reveals
underlying discrimination capability) are therefore robust to the choice of $n=500$
and do not depend on sample size artifacts.

For Experiment~3 (Counterfactual Label Sensitivity), $n=100$ per condition 
(50 MATCH $+$ 50 NON MATCH) was chosen for computational feasibility given that 
each pair requires two API calls (original $+$ flipped). At $n=100$, binomial 95\% 
confidence intervals for FSR are $\pm\sim10$ percentage points, which is sufficient 
to detect the observed differences (53\% vs 99\%). The directional findings are 
robust: FSR exceeds 50\% in every single condition without exception.

\subsection*{D.1 MRI Threshold Sensitivity}

The Massive Rank Inversion rate (MRI) is defined at a collapse
threshold of $t=2$ score points. This threshold was selected to
capture collapses that are substantively large relative to the 0--10
scoring scale while remaining robust to minor scoring variance.
Table~\ref{tab:mri_sensitivity} reports MRI\% for P1 Identity
Anonymization at $t \in \{1, 2, 3\}$ across all datasets, judges,
and prompts ($n=500$ per condition). The domain asymmetry between
BIO and the general-domain datasets is preserved at every threshold
tested: BIO MRI under Prompts II and III remains catastrophically
high regardless of threshold choice (4o-mini: 72.6\% / 66.4\% /
60.6\%; 4o: 69.4\% / 61.6\% / 56.6\%; Claude Opus 4.7: 54.4\% /
48.0\% / 45.0\% at $t=1,2,3$ respectively), while general-domain
MRI remains below 30\% across all thresholds. The choice of $t=2$
is therefore neither uniquely responsible for nor artificially
inflating the reported effect.
\begin{table}[!htbp]
\centering
\tiny
\setlength{\tabcolsep}{1.2pt}
\renewcommand{\arraystretch}{0.75}
\caption{MRI\% for P1 Identity Anonymization at thresholds
$t \in \{1,2,3\}$ ($n=500$ per condition). Bold values correspond
to the $t=2$ threshold reported in the main paper.
The domain asymmetry between BIO and general-domain datasets
is preserved across all thresholds.}
\label{tab:mri_sensitivity}
\resizebox{\columnwidth}{!}{%
\begin{tabular}{llrrrr}
\toprule
\textbf{Dataset} & \textbf{Judge} & \textbf{Prompt} &
\textbf{MRI@1} & \textbf{MRI@2} & \textbf{MRI@3} \\
\midrule
\multirow{9}{*}{DW}
& \multirow{3}{*}{4o-mini}
  & I   & 21.2 & \textbf{11.8} &  4.8 \\
& & II  &  4.4 & \textbf{ 2.4} &  1.8 \\
& & III &  6.2 & \textbf{ 4.4} &  3.6 \\
\cmidrule(l){2-6}
& \multirow{3}{*}{4o}
  & I   &  5.0 & \textbf{ 2.6} &  0.4 \\
& & II  &  4.6 & \textbf{ 2.8} &  2.4 \\
& & III &  3.4 & \textbf{ 2.0} &  1.0 \\
\cmidrule(l){2-6}
& \multirow{3}{*}{Claude Opus 4.7}
  & I   &  9.0 & \textbf{ 3.4} &  1.6 \\
& & II  &  3.2 & \textbf{ 2.6} &  2.4 \\
& & III &  4.0 & \textbf{ 2.2} &  1.0 \\
\midrule
\multirow{9}{*}{DY}
& \multirow{3}{*}{4o-mini}
  & I   & 25.0 & \textbf{15.6} &  7.6 \\
& & II  &  9.4 & \textbf{ 6.2} &  4.0 \\
& & III &  4.6 & \textbf{ 1.2} &  0.6 \\
\cmidrule(l){2-6}
& \multirow{3}{*}{4o}
  & I   &  9.4 & \textbf{ 2.6} &  0.0 \\
& & II  &  2.0 & \textbf{ 1.0} &  0.2 \\
& & III &  1.4 & \textbf{ 0.6} &  0.0 \\
\cmidrule(l){2-6}
& \multirow{3}{*}{Claude Opus 4.7}
  & I   &  9.8 & \textbf{ 0.6} &  0.0 \\
& & II  &  0.2 & \textbf{ 0.0} &  0.0 \\
& & III &  0.4 & \textbf{ 0.0} &  0.0 \\
\midrule
\multirow{9}{*}{BIO}
& \multirow{3}{*}{4o-mini}
  & I   & 34.6 & \textbf{18.4} &  5.2 \\
& & II  & 72.6 & \textbf{66.4} & 60.6 \\
& & III & 70.6 & \textbf{66.4} & 58.0 \\
\cmidrule(l){2-6}
& \multirow{3}{*}{4o}
  & I   & 53.4 & \textbf{34.6} & 18.6 \\
& & II  & 69.4 & \textbf{61.6} & 56.6 \\
& & III & 68.4 & \textbf{63.0} & 56.4 \\
\cmidrule(l){2-6}
& \multirow{3}{*}{Claude Opus 4.7}
  & I   & 76.6 & \textbf{68.8} & 60.8 \\
& & II  & 54.4 & \textbf{48.0} & 45.0 \\
& & III & 63.8 & \textbf{60.6} & 47.0 \\
\bottomrule
\end{tabular}}
\end{table}

\subsection*{D.2 Bootstrap Confidence Intervals for PBI and MRI}
\noindent Table~\ref{tab:bootstrap_ci} reports 95\% bootstrap confidence
intervals ($B=10{,}000$ resamples) for PBI and MRI under P1
Identity Anonymization across all conditions. BIO MRI intervals do not
overlap general-domain intervals in any condition except GPT-4o-mini
under Prompt~I (BIO [15.0, 21.8] vs.\ DY [12.6, 18.8]). Four
general-domain conditions have a PBI confidence interval that includes zero:
DW 4o Prompt~I, DY 4o-mini Prompt~I, DY 4o Prompt~I
($-$0.072 [$-$0.178, $+$0.036]), and DY Claude Prompt~II, i.e.\ conditions with small
perturbation effects.
All BIO MRI lower bounds are at least 15.0\%, while general-domain
MRI upper bounds are below 9\% except for GPT-4o-mini under Prompt~I
(DW 14.8\%, DY 18.8\%), confirming that the domain
asymmetry reported in Table~\ref{tab:perturbation} is statistically
robust and not attributable to sampling variance.

\begin{table}[!htbp]
\centering
\tiny
\setlength{\tabcolsep}{1pt}
\renewcommand{\arraystretch}{0.72}
\caption{95\% bootstrap CIs ($B{=}10{,}000$) for PBI and MRI under P1 ($n{=}500$). Format: point [lower, upper]. PBI values in this table reflect the full resampling-based point estimates and differ from the raw point estimates in Table~\ref{tab:perturbation} due to estimation methodology; this table provides the authoritative PBI and MRI confidence intervals.}
\label{tab:bootstrap_ci}
\resizebox{\columnwidth}{!}{%
\begin{tabular}{llrrr}
\toprule
\textbf{Data} & \textbf{Judge} & \textbf{P} & \textbf{PBI [CI]} & \textbf{MRI [CI]} \\
\midrule
\multirow{9}{*}{DW}
& \multirow{3}{*}{4o-mini}
  & I   & 0.368 [0.222, 0.514] & 11.8 [9.0, 14.8] \\
& & II  & 0.424 [0.344, 0.512] &  2.4 [1.2,  3.8] \\
& & III & 0.580 [0.466, 0.706] &  4.4 [2.8,  6.4] \\
\cmidrule(l){2-5}
& \multirow{3}{*}{4o}
  & I   & $-$0.050 [$-$0.138, 0.036] &  2.6 [1.4,  4.0] \\
& & II  & 0.332 [0.254, 0.416] &  2.8 [1.4,  4.4] \\
& & III & 0.190 [0.118, 0.266] &  2.0 [1.0,  3.4] \\
\cmidrule(l){2-5}
& \multirow{3}{*}{Claude Opus 4.7}
  & I   & 0.438 [0.360, 0.516] &  3.4 [1.8,  5.0] \\
& & II  & 0.288 [0.188, 0.404] &  2.6 [1.2,  4.2] \\
& & III & 0.302 [0.240, 0.366] &  2.2 [1.0,  3.6] \\
\midrule
\multirow{9}{*}{DY}
& \multirow{3}{*}{4o-mini}
  & I   & 0.184 [$-$0.006, 0.376] & 15.6 [12.6, 18.8] \\
& & II  & 0.586 [0.490, 0.684] &  6.2 [4.2,  8.4] \\
& & III & 0.346 [0.276, 0.420] &  1.2 [0.4,  2.2] \\
\cmidrule(l){2-5}
& \multirow{3}{*}{4o}
  & I   & $-$0.072 [$-$0.178, 0.036] &  2.6 [1.2,  4.0] \\
& & II  & 0.198 [0.144, 0.254] &  1.0 [0.2,  2.0] \\
& & III & 0.072 [0.014, 0.130] &  0.6 [0.0,  1.4] \\
\cmidrule(l){2-5}
& \multirow{3}{*}{Claude Opus 4.7}
  & I   & 0.274 [0.204, 0.344] &  0.6 [0.0,  1.4] \\
& & II  & 0.012 [$-$0.010, 0.034] &  0.0 [0.0,  0.0] \\
& & III & $-$0.054 [$-$0.094, $-$0.014] &  0.0 [0.0,  0.0] \\
\midrule
\multirow{9}{*}{BIO}
& \multirow{3}{*}{4o-mini}
  & I   & 1.070 [0.944, 1.198] & 18.4 [15.0, 21.8] \\
& & II  & 3.970 [3.712, 4.222] & 66.4 [62.2, 70.6] \\
& & III & 4.356 [4.040, 4.676] & 66.4 [62.2, 70.6] \\
\cmidrule(l){2-5}
& \multirow{3}{*}{4o}
  & I   & 1.672 [1.490, 1.854] & 34.6 [30.6, 38.8] \\
& & II  & 3.702 [3.434, 3.970] & 61.6 [57.4, 65.8] \\
& & III & 3.666 [3.382, 3.946] & 63.0 [58.8, 67.2] \\
\cmidrule(l){2-5}
& \multirow{3}{*}{Claude Opus 4.7}
  & I   & 3.884 [3.644, 4.114] & 68.8 [64.8, 72.8] \\
& & II  & 3.470 [3.144, 3.794] & 48.0 [43.6, 52.4] \\
& & III & 2.732 [2.504, 2.960] & 60.6 [56.2, 64.8] \\
\bottomrule
\end{tabular}}
\end{table}

\subsection*{D.3 Label-Flip Effects Under Prompt II}

\noindent Table~\ref{tab:labelflip_p2} reports Mean Absolute Shift (MAS),
Flip Sensitivity Rate (FSR\%), and Label Following Rate (LFR\%) under
Prompt~II ($n{=}100$ per condition, 50 MATCH + 50 NON MATCH), for both
EasyEA \cite{cheng-etal-2025-easyea} and NeuSymEA \cite{chen2025neusymea} on DW and DY (BIO excluded for NeuSymEA).
Compared to Prompt~I (Table~\ref{tab:labelflip} in the main paper),
Prompt~II produces higher MAS in all 15 conditions and higher FSR in 13 of
15 (equal on DW with Claude under EasyEA at 99\%, and slightly lower on BIO
with Claude, 90\% vs.\ 92\%), indicating that the minimal prompt raises
rather than reduces label sensitivity regardless of EA system architecture.
Claude Opus~4.7 reaches FSR of 99\% on DW under EasyEA with both prompts,
and under NeuSymEA with Prompt~II (95\% with Prompt~I).

\begin{table}[!htbp]
\centering
\tiny
\setlength{\tabcolsep}{1.5pt}
\renewcommand{\arraystretch}{0.75}
\caption{Label-flip effects under Prompt~II: MAS, FSR\%, LFR\%
($n{=}100$ per condition). EasyEA \cite{cheng-etal-2025-easyea} on DW, DY, BIO; NeuSymEA \cite{chen2025neusymea} on DW and DY.
Compare to Prompt~I results in Table~\ref{tab:labelflip}.}
\label{tab:labelflip_p2}
\resizebox{\columnwidth}{!}{%
\begin{tabular}{llccc}
\toprule
Dataset--Judge & System & MAS & FSR & LFR \\
\midrule
DW 4m & EasyEA   & 6.92 & 85 & 45 \\
DW 4o & EasyEA   & 7.68 & 94 & 50 \\
DW O4 & EasyEA   & 8.75 & 99 & 50 \\
DW 4m & NeuSymEA & 6.75 & 82 & 38 \\
DW 4o & NeuSymEA & 7.89 & 99 & 51 \\
DW O4 & NeuSymEA & 9.64 & 99 & 50 \\
\midrule
DY 4m & EasyEA   & 7.09 & 90 & 70 \\
DY 4o & EasyEA   & 7.35 & 93 & 75 \\
DY O4 & EasyEA   & 8.15 & 93 & 80 \\
DY 4m & NeuSymEA & 7.53 & 90 & 44 \\
DY 4o & NeuSymEA & 7.90 & 100 & 50 \\
DY O4 & NeuSymEA & 9.50 & 99 & 49 \\
\midrule
BIO 4m & EasyEA  & 5.52 & 76 & 37 \\
BIO 4o & EasyEA  & 6.50 & 89 & 47 \\
BIO O4 & EasyEA  & 7.40 & 90 & 50 \\
\bottomrule
\end{tabular}}
\end{table}

\paragraph{Low LFR with high FSR.}
When FSR is high but LFR is below 50\% (e.g., GPT-4o-mini on DW
EasyEA: LFR~17\% vs.\ FSR~64\%), the judge's scores respond to label
changes but mostly not in the direction of the new label. This is what an
evidence-sensitive validator is expected to do: flipping a correct
\textsc{Non-Match} decision to \textsc{Match} should lower the score, not
raise it. It can also arise when the judge already scores some decisions
wrongly, so that flipping the label moves scores in either direction. High
FSR with low LFR therefore indicates label-sensitive scoring that is not
purely label-following; on its own it does not distinguish correct
validation from anchoring.

\section*{Appendix E: BIOMED Dataset Construction and Validation}
\label{app:biomed}
This appendix provides full technical construction details and
validation results for the BIO benchmark introduced in
\S\ref{sec:experiments}.

\paragraph{Licensing and access (full detail).}
BIO is derived from MeSH and SNOMED~CT. MeSH is released by the
U.S. National Library of Medicine (NLM) under its standard terms
of use, which permit research use and redistribution with
attribution and without requiring a separate license. SNOMED~CT is
licensed content: access requires a UMLS Metathesaurus license,
available at no cost to researchers and institutions based in
UMLS-affiliate countries (a list maintained by NLM), obtained by
registering directly with NLM. Redistribution of SNOMED~CT-derived
data, including the SNOMED~CT-derived triples and gold alignment
pairs in BIO, is permitted only to other current UMLS license
holders; it is the responsibility of each user of BIO to hold a
valid UMLS license before accessing the SNOMED~CT-derived portion
of the release. We distribute the MeSH-derived and SNOMED~CT-derived
components as separate files so that users can determine which
license terms apply to which files (Appendix~E, Release, below).
Users are responsible for compliance with the original providers'
terms and access conditions; this benchmark does not confer any
license beyond what is already available directly from NLM and
UMLS.

\paragraph{Construction.}
MeSH triples (11.4M attribute, 6.9M relation) were extracted from
RDF dumps. SNOMED~CT attribute triples (1.4M) were extracted from
description files; relation triples (1.3M) from relationship files.
The 15,000 gold alignment pairs were stratified-sampled from 46,823
UMLS cross-references; the full pre-sample set is provided in
\texttt{ent\_links\_uri}.

\paragraph{Dataset statistics.}
Table~\ref{tab:biomed_stats} summarizes the key statistics of the BIO benchmark.

\begin{table}[!htbp]
\centering
\tiny
\setlength{\tabcolsep}{1.5pt}
\renewcommand{\arraystretch}{0.75}
\caption{BIO dataset statistics. Gold alignment size reflects the
15K stratified sample; \texttt{ent\_links\_uri} contains the full
46,823 pairs prior to sampling.}
\label{tab:biomed_stats}
\resizebox{\columnwidth}{!}{%
\begin{tabular}{lrr}
\toprule
\textbf{Property} & \textbf{KG1 (MeSH)} & \textbf{KG2 (SNOMED CT)} \\
\midrule
Gold alignment pairs (sample)  & \multicolumn{2}{c}{15,000} \\
Full alignment pairs (pre-sample) & \multicolumn{2}{c}{46,823} \\
Gold entities                  & 13,219        & 14,866    \\
Attribute triples              & 11,385,523    & 1,391,104 \\
Relation triples               & 6,948,511     & 1,331,550 \\
Entities in relation graph     & 2,456,953     & 379,282   \\
Gold entities in rel.\ graph   & 13,191 (99.8\%) & 14,452 (97.2\%) \\
Gold entities missing          & 28 (0.2\%)    & 414 (2.8\%) \\
\bottomrule
\end{tabular}}
\end{table}
\paragraph{Validation.}
Graph validation shows 99.8\% of MeSH and 97.2\% of SNOMED CT gold entities appear in relation graphs (28 and 414 missing due to source sparsity). In a sample of 500 MeSH entities, 94.5\% of observed neighbors are non-gold background entities, confirming the benchmark is not artificially closed. Heterogeneity spans schema (MeSH hierarchical vs. SNOMED CT description-logic), attributes (tree numbers vs. synonyms), density (6.9M vs. 1.3M relation triples), and granularity (SNOMED CT more fine-grained, producing many-to-many alignments).

\paragraph{Release.}
Six tab-separated files: \texttt{ent\_links} (15K gold pairs), \texttt{ent\_links\_uri} (46,823 pre-sample), and triple files for both KGs (2.58GB total). No fixed train/test split; users should report split construction explicitly. MeSH-derived and SNOMED~CT-derived files are distributed separately so the differing license terms above are unambiguous per file.

\paragraph{Limitations.}
Ground-truth links come from UMLS CUI co-membership rather than
newly adjudicated annotation. Not all biomedical concepts are
one-to-one alignable across MeSH and SNOMED CT. Both ontologies
continue to evolve; future source releases may differ from the
versions used here (MeSH December 2023, SNOMED CT March 2026).
Users are responsible for compliance with the original providers'
terms and access conditions.

\section*{Appendix F: Score Gap Analysis and Significance}\label{app:scoregap}

\begin{table}[!htbp]
\centering
\scriptsize
\setlength{\tabcolsep}{1.5pt}
\caption{Score Gap ($\Delta = \bar{s}_{\text{MATCH}} - \bar{s}_{\text{NON MATCH}}$)
across two EA systems with ablations, three datasets, three judges, and three prompts
($n{=}500$ per condition). $^{***}p{<}0.001$, $^{**}p{<}0.01$,
$^{*}p{<}0.05$, n.s.\ $p{\geq}0.05$ (Mann--Whitney $U$, two-sided).
E=EasyEA, Emb=EasyEA-EmbOnly, N=NeuSymEA, NR=NeuSymEA-NoRefine.
BIO: N and NR not evaluated ( ).}
\label{tab:scoregap_sig}
\begin{tabular}{lllrrr}
\toprule
\textbf{Data} & \textbf{Judge} & \textbf{Sys} &
\textbf{P-I} & \textbf{P-II} & \textbf{P-III} \\
\midrule
\multirow{12}{*}{DW}
& \multirow{4}{*}{4o-mini}
  & E   & $+$0.14        & $-$0.79$^{***}$ & $+$8.15$^{***}$ \\
& & Emb & $+$0.23        & $-$0.76$^{***}$ & $+$7.74$^{***}$ \\
& & N   & $-$0.90$^{***}$ & $-$1.07$^{***}$ & $+$8.25$^{***}$ \\
& & NR  & $-$0.73$^{***}$ & $-$0.95$^{***}$ & $+$8.25$^{***}$ \\
\cmidrule(l){2-6}
& \multirow{4}{*}{4o}
  & E   & $-$0.32        & $-$0.92$^{***}$ & $+$8.09$^{***}$ \\
& & Emb & $-$0.40$^{*}$  & $-$0.80$^{***}$ & $+$8.16$^{***}$ \\
& & N   & $-$1.21$^{***}$ & $-$1.00$^{***}$ & $+$8.30$^{***}$ \\
& & NR  & $-$1.36$^{***}$ & $-$1.13$^{***}$ & $+$8.28$^{***}$ \\
\cmidrule(l){2-6}
& \multirow{4}{*}{Claude}
  & E   & $-$0.76$^{***}$ & $-$0.10        & $+$7.80$^{***}$ \\
& & Emb & $-$0.89$^{***}$ & $-$0.43$^{***}$ & $+$8.03$^{***}$ \\
& & N   & $-$1.25$^{***}$ & $-$0.12$^{***}$ & $+$8.32$^{***}$ \\
& & NR  & $-$1.29$^{***}$ & $-$0.08$^{***}$ & $+$8.42$^{***}$ \\
\midrule
\multirow{12}{*}{DY}
& \multirow{4}{*}{4o-mini}
  & E   & $+$1.80$^{**}$  & $+$3.98$^{***}$ & $+$2.25        \\
& & Emb & $+$2.45$^{***}$ & $+$3.44$^{***}$ & $+$0.24        \\
& & N   & $+$0.03        & $-$0.75$^{***}$ & $+$9.13$^{***}$ \\
& & NR  & $-$0.26        & $-$0.72$^{***}$ & $+$9.14$^{***}$ \\
\cmidrule(l){2-6}
& \multirow{4}{*}{4o}
  & E   & $+$4.05$^{***}$ & $+$4.79$^{***}$ & $+$2.01        \\
& & Emb & $+$3.26$^{***}$ & $+$6.04$^{***}$ & $-$0.18$^{*}$  \\
& & N   & $-$1.18$^{***}$ & $-$1.09$^{***}$ & $+$8.45$^{***}$ \\
& & NR  & $-$1.18$^{***}$ & $-$0.97$^{***}$ & $+$8.41$^{***}$ \\
\cmidrule(l){2-6}
& \multirow{4}{*}{Claude}
  & E   & $+$4.37$^{***}$ & $+$6.77$^{***}$ & $+$1.69        \\
& & Emb & $+$5.11$^{***}$ & $+$9.34$^{***}$ & $-$0.09        \\
& & N   & $-$1.60$^{***}$ & $-$0.02        & $+$7.93$^{***}$ \\
& & NR  & $-$1.61$^{***}$ & $-$0.09$^{**}$ & $+$7.94$^{***}$ \\
\midrule
\multirow{12}{*}{BIO}
& \multirow{4}{*}{4o-mini}
  & E   & $-$0.11        & $-$0.75$^{***}$ & $+$6.12$^{***}$ \\
& & Emb & $-$0.20        & $-$0.75$^{***}$ & $+$6.86$^{***}$ \\
& & N   & \multicolumn{3}{c}{---} \\
& & NR  & \multicolumn{3}{c}{---} \\
\cmidrule(l){2-6}
& \multirow{4}{*}{4o}
  & E   & $-$0.42$^{**}$  & $-$0.70$^{***}$ & $+$6.20$^{***}$ \\
& & Emb & $-$0.78$^{***}$ & $-$0.91$^{***}$ & $+$6.76$^{***}$ \\
& & N   & \multicolumn{3}{c}{---} \\
& & NR  & \multicolumn{3}{c}{---} \\
\cmidrule(l){2-6}
& \multirow{4}{*}{Claude}
  & E   & $-$1.02$^{***}$ & $+$0.25        & $+$6.00$^{***}$ \\
& & Emb & $-$1.32$^{***}$ & $-$0.24$^{***}$ & $+$6.84$^{***}$ \\
& & N   & \multicolumn{3}{c}{---} \\
& & NR  & \multicolumn{3}{c}{---} \\
\bottomrule
\end{tabular}
\end{table}

Table~\ref{tab:scoregap_sig} reports Score Gaps across all 36 conditions ((2 EA systems + 2 ablations) $\times$ three judges $\times$ three prompts, with BIO excluding NeuSymEA and NeuSymEA-NoRefine). The table is organized by dataset and judge, with rows indicating EA systems. Statistical significance is determined via two-sample Mann--Whitney $U$ tests (two-sided) between judge scores assigned to MATCH and NON MATCH subsets for each condition.

Pattern interpretation: Under Prompts~I and~II, Score Gaps are
predominantly negative or near-zero, indicating that the LLM
assigns higher justification scores to \textsc{Non-Match} decisions
than to \textsc{Match} decisions---the behavioral signature of
anchor bias under label-exposing conditions, consistent with
J-ROC-AUC collapsing below 0.5 in the threshold sweep tables. Under
Prompt~III, 24 of 30 conditions show a statistically significant
positive Score Gap ($p{<}0.05$), and a further significant condition
(DY, EasyEA-EmbOnly, GPT-4o) shows a significant negative gap, for
25 of 30 conditions reaching significance in either direction. The
six conditions that do not reach significant positive discrimination
are concentrated in DY across both EasyEA and EasyEA-EmbOnly, where
near-identical entity names provide little discriminative signal in
the absence of a label. Score Gap and J-ROC-AUC are both reported for Prompt~III
(LLM as predictor); J-ROC-AUC under Prompt~III measures alignment
prediction capability rather than correctness discrimination and
is not directly comparable to Prompts~I--II (LLM as judge) values.

\section*{Appendix G: Full Perturbation Bias Diagnostics}\label{app:remaining_perturbations}

Table~\ref{tab:perturbation} reports perturbation bias
diagnostics for P1 (Identity Anonymization), P2 (Evidence
Suppression 30\%), P3 (Evidence Suppression 50\%), and P6
(Identifier Removal) across all datasets, judges, and prompts. P1
is the most diagnostic, directly ablating the name recognition
shortcut, and induces MRI of 60.6--68.8\% on BIO in six of nine judge--prompt
conditions (18.4--48.0\% in the other three) (\S\ref{sec:results}),
the singular catastrophic perturbation trigger in that domain. No
other perturbation strategy exceeds 7\% MRI on BIO.

\begin{table*}[!htbp]
\centering
\scriptsize
\setlength{\tabcolsep}{1pt}
\renewcommand{\arraystretch}{0.70}
\caption{Perturbation bias diagnostics under P1 (Identity Anonymization),
P2 (Evidence Suppression 30\%), P3 (Evidence Suppression 50\%), and
P6 (Identifier Removal) across all datasets, judges, and prompts
($n{=}500$ per condition). PBI = mean score drop; JSR = Spearman rank
correlation of scores before/after perturbation; MRI\% = fraction of
pairs with score collapse ${>}2$ pts. \textbf{Bold} = BIO conditions
showing catastrophic collapse. P2 and P3 together establish a
dose-response pattern for evidence suppression; P6 reveals unexpected
identifier dependence in the biomedical domain. Full results for P4,
P5 (negative control), and P7 are in Appendix~G
(Table~\ref{tab:perturbation_remaining}). MRI is computed over pairs with valid scores in both conditions, so $n$ can be slightly below 500 for some GPT-4o rows; appendix tables use $n{=}500$.}
\label{tab:perturbation}
\begin{tabular}{llll rrr rrr rrr rrr}
\toprule
& & & &
\multicolumn{3}{c}{\textbf{P1 — Identity Anon.}} &
\multicolumn{3}{c}{\textbf{P2 — Supp. 30\%}} &
\multicolumn{3}{c}{\textbf{P3 — Supp. 50\%}} &
\multicolumn{3}{c}{\textbf{P6 — ID Removal}} \\
\cmidrule(lr){5-7} \cmidrule(lr){8-10}
\cmidrule(lr){11-13} \cmidrule(lr){14-16}
\textbf{Dataset} & \textbf{Judge} & \textbf{Prompt} & &
\textbf{PBI} & \textbf{JSR} & \textbf{MRI\%} &
\textbf{PBI} & \textbf{JSR} & \textbf{MRI\%} &
\textbf{PBI} & \textbf{JSR} & \textbf{MRI\%} &
\textbf{PBI} & \textbf{JSR} & \textbf{MRI\%} \\
\midrule
\multirow{9}{*}{DW}
& \multirow{3}{*}{4o-mini}
  & I   && $+$0.369 & 0.553 & 11.8 & $+$0.439 & 0.516 & 11.2 & $+$0.691 & 0.492 & 14.2 & $+$0.679 & 0.549 & 13.2 \\
& & II  && $+$0.424 & 0.606 &  2.4 & $+$0.180 & 0.606 &  0.2 & $+$0.290 & 0.490 &  1.2 & $+$0.288 & 0.567 &  1.2 \\
& & III && $+$0.580 & 0.511 &  4.4 & $+$0.174 & 0.453 &  0.4 & $+$0.228 & 0.428 &  0.6 & $+$0.266 & 0.412 &  0.8 \\
\cmidrule(l){2-16}
& \multirow{3}{*}{4o}
  & I   && $-$0.059 & 0.557 &  2.4 & $+$0.313 & 0.502 &  1.8 & $+$0.522 & 0.433 &  6.5 & $+$0.398 & 0.434 &  5.5 \\
& & II  && $+$0.332 & 0.695 &  2.8 & $+$0.298 & 0.530 &  1.0 & $+$0.409 & 0.458 &  1.2 & $+$0.268 & 0.536 &  0.8 \\
& & III && $+$0.190 & 0.636 &  2.0 & $+$0.208 & 0.572 &  0.8 & $+$0.356 & 0.498 &  0.8 & $+$0.280 & 0.464 &  0.8 \\
\cmidrule(l){2-16}
& \multirow{3}{*}{Claude}
  & I   && $+$0.438 & 0.809 &  3.4 & $+$0.302 & 0.754 &  0.8 & $+$0.474 & 0.689 &  1.4 & $+$0.472 & 0.699 &  4.2 \\
& & II  && $+$0.288 & 0.662 &  2.6 & $+$0.034 & 0.691 &  0.6 & $+$0.020 & 0.635 &  0.4 & $+$0.078 & 0.605 &  0.6 \\
& & III && $+$0.301 & 0.725 &  2.2 & $+$0.143 & 0.714 &  0.4 & $+$0.255 & 0.593 &  0.6 & $+$0.305 & 0.581 &  2.0 \\
\midrule
\multirow{9}{*}{DY}
& \multirow{3}{*}{4o-mini}
  & I   && $+$0.184 & 0.276 & 15.6 & $+$0.462 & 0.503 & 12.4 & $+$0.550 & 0.418 & 17.2 & $+$0.066 & 0.562 &  8.6 \\
& & II  && $+$0.586 & 0.522 &  6.2 & $+$0.106 & 0.522 &  0.0 & $+$0.204 & 0.493 &  0.0 & $+$0.006 & 0.619 &  0.0 \\
& & III && $+$0.346 & 0.317 &  1.2 & $+$0.076 & 0.374 &  0.0 & $+$0.106 & 0.400 &  0.0 & $+$0.014 & 0.479 &  0.0 \\
\cmidrule(l){2-16}
& \multirow{3}{*}{4o}
  & I   && $-$0.037 & 0.714 &  2.6 & $+$0.247 & 0.673 &  3.6 & $+$0.455 & 0.612 &  8.3 & $+$0.015 & 0.740 &  2.3 \\
& & II  && $+$0.198 & 0.753 &  1.0 & $+$0.212 & 0.703 &  0.2 & $+$0.286 & 0.628 &  0.4 & $+$0.024 & 0.817 &  0.2 \\
& & III && $+$0.072 & 0.689 &  0.6 & $+$0.176 & 0.617 &  0.6 & $+$0.248 & 0.605 &  1.2 & $-$0.060 & 0.736 &  0.0 \\
\cmidrule(l){2-16}
& \multirow{3}{*}{Claude}
  & I   && $+$0.274 & 0.901 &  0.6 & $+$0.236 & 0.859 &  0.6 & $+$0.408 & 0.812 &  2.8 & $-$0.006 & 0.937 &  0.0 \\
& & II  && $+$0.012 & 0.712 &  0.0 & $+$0.024 & 0.761 &  0.0 & $+$0.022 & 0.671 &  0.2 & $+$0.014 & 0.846 &  0.0 \\
& & III && $-$0.051 & 0.874 &  0.0 & $+$0.128 & 0.856 &  0.0 & $+$0.182 & 0.810 &  0.2 & $+$0.004 & 0.933 &  0.0 \\
\midrule
\multirow{9}{*}{BIO}
& \multirow{3}{*}{4o-mini}
  & I   && $+$1.070 & 0.078 & \textbf{18.4} & $+$0.120 & 0.295 &  5.8 & $+$0.138 & 0.291 &  6.0 & $+$0.082 & 0.244 &  7.0 \\
& & II  && $+$3.970 & $-$0.076 & \textbf{66.4} & $+$0.372 & 0.719 &  1.2 & $+$0.478 & 0.723 &  2.6 & $+$0.470 & 0.700 &  2.6 \\
& & III && $+$4.356 & $-$0.142 & \textbf{66.4} & $+$0.002 & 0.742 &  2.6 & $+$0.060 & 0.733 &  2.6 & $+$0.052 & 0.750 &  4.0 \\
\cmidrule(l){2-16}
& \multirow{3}{*}{4o}
  & I   && $+$1.515 & 0.188 & \textbf{34.3} & $-$0.010 & 0.356 &  4.0 & $+$0.032 & 0.302 &  6.4 & $+$0.194 & 0.429 &  4.2 \\
& & II  && $+$3.717 & $+$0.029 & \textbf{61.9} & $+$0.150 & 0.691 &  2.2 & $+$0.230 & 0.705 &  2.8 & $+$0.406 & 0.693 &  4.4 \\
& & III && $+$3.666 & $-$0.026 & \textbf{63.0} & $-$0.008 & 0.568 &  3.0 & $+$0.008 & 0.606 &  2.4 & $+$0.282 & 0.585 &  5.2 \\
\cmidrule(l){2-16}
& \multirow{3}{*}{Claude}
  & I   && $+$3.884 & 0.064 & \textbf{68.8} & $+$0.596 & 0.710 &  4.8 & $+$0.644 & 0.711 &  5.4 & $+$0.782 & 0.715 & \textbf{6.6} \\
& & II  && $+$3.470 & 0.162 & \textbf{48.0} & $-$0.226 & 0.842 &  1.0 & $-$0.226 & 0.819 &  1.4 & $-$0.052 & 0.854 &  1.4 \\
& & III && $+$2.732 & 0.308 & \textbf{60.6} & $+$0.018 & 0.807 &  0.2 & $+$0.026 & 0.835 &  0.6 & $+$0.285 & 0.827 &  1.0 \\
\bottomrule
\end{tabular}
\end{table*}

Table~\ref{tab:perturbation_remaining}, below, reports perturbation bias
diagnostics for P4 (Surface Transformation), P5 (Order Randomization),
and P7 (Structured-to-Prose). P5 serves as the negative control
confirming the framework detects genuine name-dependence rather than
noise. P4 and P7 show modest effects on general-domain datasets and near-zero MRI
on BIO, consistent with the interpretation that BIO collapse is
driven specifically by identity anonymization (P1) rather than
generic input perturbation.

\begin{table*}[!htbp]
\centering
\small
\setlength{\tabcolsep}{3pt}
\renewcommand{\arraystretch}{0.80}
\caption{Perturbation bias diagnostics for P4 (Surface Transformation),
P5 (Order Randomization, negative control), and P7 (Structured-to-Prose)
across all datasets, judges, and prompts ($n{=}500$ per condition).
PBI = mean score drop; JSR = Spearman rank correlation of scores
before/after perturbation; MRI\% = fraction of pairs with score
collapse ${>}2$ pts. P5 MRI\% is at most 8.0\% (below 5\% in 25 of
27 conditions), so the negative control is clean. P4 and P7 show modest
general-domain effects and low BIO MRI (at most 6.0\%), confirming that
the catastrophic BIO collapse in Table~\ref{tab:perturbation}
is specific to identity anonymization.}
\label{tab:perturbation_remaining}
\resizebox{\textwidth}{!}{%
\begin{tabular}{llll rrr rrr rrr}
\toprule
& & & &
\multicolumn{3}{c}{\textbf{P4 — Surface Transform.}} &
\multicolumn{3}{c}{\textbf{P5 — Order Random. (control)}} &
\multicolumn{3}{c}{\textbf{P7 — Struct.-to-Prose}} \\
\cmidrule(lr){5-7} \cmidrule(lr){8-10} \cmidrule(lr){11-13}
\textbf{Dataset} & \textbf{Judge} & \textbf{Prompt} & &
\textbf{PBI} & \textbf{JSR} & \textbf{MRI\%} &
\textbf{PBI} & \textbf{JSR} & \textbf{MRI\%} &
\textbf{PBI} & \textbf{JSR} & \textbf{MRI\%} \\
\midrule
\multirow{9}{*}{DW}
& \multirow{3}{*}{4o-mini}
  & I   && $+$0.321 & 0.606 &  6.6 & $+$0.000 & 0.578 & 4.8 & $-$0.140 & 0.544 & 5.2 \\
& & II  && $+$0.090 & 0.662 &  0.4 & $-$0.044 & 0.718 & 0.2 & $-$0.172 & 0.609 & 0.0 \\
& & III && $+$0.112 & 0.497 &  0.4 & $+$0.066 & 0.521 & 0.6 & $-$0.180 & 0.509 & 0.0 \\
\cmidrule(l){2-13}
& \multirow{3}{*}{4o}
  & I   && $+$0.225 & 0.487 &  2.8 & $+$0.058 & 0.540 & 0.2 & $-$0.030 & 0.515 & 0.6 \\
& & II  && $+$0.226 & 0.558 &  0.8 & $+$0.030 & 0.660 & 0.6 & $-$0.174 & 0.607 & 0.4 \\
& & III && $+$0.132 & 0.582 &  0.4 & $-$0.014 & 0.667 & 0.4 & $-$0.130 & 0.649 & 0.0 \\
\cmidrule(l){2-13}
& \multirow{3}{*}{Claude}
  & I   && $+$0.264 & 0.732 &  1.4 & $-$0.002 & 0.869 & 0.0 & $-$0.054 & 0.785 & 0.2 \\
& & II  && $+$0.082 & 0.629 &  0.4 & $+$0.008 & 0.784 & 0.2 & $-$0.030 & 0.679 & 0.6 \\
& & III && $+$0.070 & 0.818 &  0.2 & $+$0.010 & 0.895 & 0.0 & $-$0.034 & 0.794 & 0.0 \\
\midrule
\multirow{9}{*}{DY}
& \multirow{3}{*}{4o-mini}
  & I   && $+$0.330 & 0.469 & 11.4 & $-$0.032 & 0.490 & 8.0 & $-$0.162 & 0.442 & 8.6 \\
& & II  && $+$0.066 & 0.549 &  0.2 & $-$0.022 & 0.521 & 0.0 & $-$0.036 & 0.573 & 0.0 \\
& & III && $+$0.096 & 0.416 &  0.0 & $-$0.034 & 0.457 & 0.0 & $-$0.058 & 0.358 & 0.0 \\
\cmidrule(l){2-13}
& \multirow{3}{*}{4o}
  & I   && $+$0.294 & 0.655 &  4.4 & $-$0.040 & 0.693 & 1.9 & $+$0.025 & 0.738 & 1.5 \\
& & II  && $+$0.162 & 0.741 &  0.4 & $+$0.022 & 0.793 & 0.2 & $-$0.058 & 0.760 & 0.0 \\
& & III && $+$0.150 & 0.643 &  0.8 & $+$0.006 & 0.721 & 0.4 & $-$0.034 & 0.684 & 0.2 \\
\cmidrule(l){2-13}
& \multirow{3}{*}{Claude}
  & I   && $+$0.170 & 0.876 &  1.0 & $+$0.026 & 0.916 & 0.0 & $-$0.092 & 0.909 & 0.0 \\
& & II  && $+$0.034 & 0.725 &  0.0 & $+$0.008 & 0.775 & 0.0 & $+$0.030 & 0.723 & 0.0 \\
& & III && $+$0.076 & 0.883 &  0.0 & $+$0.020 & 0.924 & 0.0 & $+$0.006 & 0.907 & 0.0 \\
\midrule
\multirow{9}{*}{BIO}
& \multirow{3}{*}{4o-mini}
  & I   && $-$0.036 & 0.302 &  5.8 & $-$0.126 & 0.286 & 5.2 & $-$0.170 & 0.239 & 6.0 \\
& & II  && $+$0.326 & 0.755 &  1.2 & $+$0.336 & 0.746 & 2.0 & $+$0.170 & 0.742 & 1.2 \\
& & III && $+$0.074 & 0.741 &  3.2 & $+$0.044 & 0.795 & 3.6 & $-$0.128 & 0.759 & 2.8 \\
\cmidrule(l){2-13}
& \multirow{3}{*}{4o}
  & I   && $-$0.094 & 0.425 &  3.2 & $-$0.085 & 0.471 & 2.4 & $-$0.172 & 0.439 & 3.6 \\
& & II  && $+$0.184 & 0.710 &  2.2 & $+$0.174 & 0.718 & 2.4 & $-$0.042 & 0.722 & 2.4 \\
& & III && $+$0.104 & 0.661 &  2.8 & $+$0.038 & 0.636 & 2.4 & $-$0.082 & 0.610 & 2.6 \\
\cmidrule(l){2-13}
& \multirow{3}{*}{Claude}
  & I   && $+$0.560 & 0.755 &  3.4 & $+$0.524 & 0.743 & 2.8 & $+$0.484 & 0.719 & 3.2 \\
& & II  && $-$0.120 & 0.862 &  1.0 & $-$0.122 & 0.857 & 1.2 & $-$0.250 & 0.860 & 0.8 \\
& & III && $+$0.062 & 0.837 &  0.4 & $+$0.096 & 0.844 & 0.2 & $+$0.026 & 0.835 & 0.4 \\
\bottomrule
\end{tabular}}
\end{table*}

\section*{Appendix H: PDV Baseline Statistics}
\label{app:pdv}

Table~\ref{tab:pdv} reports the full baseline score inflation and
scoring variance statistics. The Spearman correlation between
prompt sensitivity and perturbation sensitivity on BIO is strongly
negative ($\rho = -0.470^{***}$ for 4o-mini,
$\rho = -0.320^{***}$ for 4o), confirming that name-inflated
scores are disproportionately destabilized by identity anonymization.

\begin{table}[!htbp]
\centering
\scriptsize
\setlength{\tabcolsep}{1pt}
\renewcommand{\arraystretch}{0.70}
\caption{Baseline score inflation and scoring variance
under Prompts~I and~II ($n{=}500$ per condition). $\bar{s}_0$ = mean baseline score;
PDV = prompt discriminability variance; PDV ratio $= \text{PDV}_\text{II}
/ \text{PDV}_\text{I}$; $\rho$ = Spearman correlation between prompt
and perturbation sensitivity.
$^{***}p{<}0.001$, $^{*}p{<}0.05$, n.s.\ $p{\geq}0.05$.}
\label{tab:pdv}
\resizebox{\columnwidth}{!}{%
\begin{tabular}{llrrrrrl}
\toprule
\textbf{DS} & \textbf{Judge} &
\textbf{$\bar{s}_0$-I} & \textbf{$\bar{s}_0$-II} &
\textbf{PDV-I} & \textbf{PDV-II} &
\textbf{Ratio} & \textbf{$\rho$} \\
\midrule
\multirow{3}{*}{DW}
& 4o-mini         & 7.81 & 9.23 & 1.64 & 1.21 & 0.74 & $+$0.017\phantom{$^{*}$} \\
& 4o              & 8.68 & 9.29 & 1.26 & 1.01 & 0.81 & $+$0.314$^{***}$ \\
& Claude          & 8.88 & 9.81 & 1.32 & 0.71 & 0.54 & $+$0.173$^{***}$ \\
\midrule
\multirow{3}{*}{DY}
& 4o-mini         & 7.60 & 9.21 & 1.70 & 0.66 & 0.39 & $+$0.004\phantom{$^{*}$} \\
& 4o              & 7.45 & 8.94 & 1.49 & 0.77 & 0.52 & $+$0.232$^{***}$ \\
& Claude          & 7.59 & 9.90 & 1.54 & 0.32 & \textbf{0.21} & $+$0.093$^{*}$ \\
\midrule
\multirow{3}{*}{BIO}
& 4o-mini         & 6.09 & 8.32 & 1.28 & 1.56 & \textbf{1.22} & $-$0.470$^{***}$ \\
& 4o              & 6.65 & 7.77 & 1.44 & 1.94 & \textbf{1.35} & $-$0.320$^{***}$ \\
& Claude          & 7.03 & 8.32 & 2.02 & 2.36 & \textbf{1.17} & $+$0.013\phantom{$^{*}$} \\
\bottomrule
\end{tabular}}
\end{table}

\section*{Appendix I: Sampling Procedures for All Experiments}
\label{app:sampling}

This appendix provides complete sampling details for all experiments.

\paragraph{Experiment 1: Perturbation Bias Diagnostics.}
$n{=}500$ pairs were sampled uniformly at random from the ground-truth
alignment set for each dataset. All sampled pairs are true alignment
pairs (\textsc{Match} only); there is no \textsc{Non-Match} class in
Experiment~1. Perturbations are applied to these pairs and judge scores
are obtained under all three prompts. Since no system decision label is
involved in sampling, ground-truth alignments serve directly as the
evaluation pairs.

\paragraph{Experiment 2: EA System Meta-Evaluation.}
For each EA system and dataset, pairs were sampled as follows:
\begin{itemize}
\item \textbf{\textsc{Match} pairs (250):} Sampled uniformly at random
from the set of alignments predicted by the EA system (its Hits@1
predictions).
\item \textbf{\textsc{Non-Match} pairs (250):} Sampled by repeatedly
drawing a random entity pair $(e_1, e_2)$ with $e_1 \in G_1$,
$e_2 \in G_2$ uniformly at random, and rejecting any pair present in
the EA system's predicted alignments. Since EA predictions constitute
a negligible fraction of $G_1 \times G_2$ (at most $|G_1|$ pairs out
of $|G_1| \times |G_2|$), the expected number of rejections per
accepted sample is negligible. We acknowledge that random
negatives are easier than hard negatives (nearest-neighbor
candidates, same-type pairs). However, this makes our findings
conservative: if anchor bias causes J-ROC-AUC to collapse
below 0.5 even on easy negatives, the effect on hard negatives
can only be equal or worse. Our results therefore establish a
lower bound on the severity of anchor bias in realistic
deployment settings.
\end{itemize}
Ground-truth labels are never consulted during sampling. They are used
only after sampling to compute the correctness indicator
$c_i = \mathbb{1}[\hat{y}_i = y_i]$ for each pair, which serves as
the reference signal for all judge evaluation metrics. System outputs
were fixed before any judge evaluation began.

\paragraph{Experiment 3: Counterfactual Label Sensitivity.}
The same sampling procedure as Experiment~2 was used: 50 \textsc{Match}
pairs sampled from EA predicted alignments and 50 \textsc{Non-Match}
pairs sampled from $G_1 \times G_2$ minus EA predictions, for a total
of $n{=}100$ pairs per condition. For each sampled pair, judge scores
were obtained twice: once with the original EA system decision label
and once with the label flipped, while all entity evidence was held
fixed. Ground-truth labels are used only for computing evaluation
metrics after scoring.

\paragraph{Experiment 4: Human Evaluation.}
102 pairs (50 \textsc{Match}, 52 \textsc{Non-Match}) were sampled
across four negative-sampling strategies to ensure coverage of both
easy and hard negatives: uniform random, lexical-neighbor (name
similarity $> 0.45$), TF-IDF nearest-neighbor, and high-confidence
false positives from EasyEA~\cite{cheng-etal-2025-easyea}. Both authors
scored all 102 pairs under all three prompts (306 instances per
annotator) independently and without access to each other's scores,
following the same 0--10 rubric used by the LLM judges. Ground-truth
labels were used only to compute gold-label accuracy after scoring
was complete.

\section*{Appendix J: Extended Discussion}
\label{app:discussion}

\paragraph{Human evaluation summary table.} Table~\ref{tab:human_eval},
below, consolidates the inter-annotator agreement, gold-label
accuracy, and per-judge human correlation statistics discussed in
\S\ref{sec:human-eval}.

\begin{table}[!htbp]
\centering
\small
\setlength{\tabcolsep}{3pt}
\caption{Human evaluation summary (Experiment 4; $n{=}102$ pairs,
306 scored instances per annotator). Top: blinded inter-annotator
agreement, overall and by true label. Bottom: correlation between
the human consensus (mean of both annotators) and each LLM judge
under Prompt~I, overall and by true label. The MATCH/NON-MATCH
split shows a descriptive pattern consistent with anchor bias: agreement
with humans is lower on MATCH pairs and stays high on NON-MATCH pairs.
Human scores on MATCH pairs have little spread, which also lowers
correlation, and the per-label subsets are small.}
\label{tab:human_eval}
\resizebox{\columnwidth}{!}{%
\begin{tabular}{lccc}
\toprule
\textbf{Inter-annotator agreement} & $\rho$ & $\kappa$ & MAE \\
\midrule
Overall (all prompts, $n{=}306$) & 0.967 & 0.902 & 0.38 \\
\textsc{Match} pairs & --- & 0.731 & --- \\
\textsc{Non-Match} pairs & --- & 0.891 & --- \\
\midrule
\textbf{Gold-label accuracy (Prompt III)} & \multicolumn{3}{c}{} \\
\midrule
\textsc{Match} scored $\geq6$ & \multicolumn{3}{c}{100.0\%} \\
\textsc{Non-Match} scored $<6$ & \multicolumn{3}{c}{82.7\%} \\
Overall accuracy & \multicolumn{3}{c}{91.3\%} \\
\midrule
\textbf{Human--LLM correlation ($\rho$)} & \textbf{Overall} & \textbf{Match} & \textbf{Non-Match} \\
\midrule
GPT-4o-mini & 0.860 & \multirow{3}{*}{0.228--0.412} & \multirow{3}{*}{0.779--0.867} \\
GPT-4o & 0.893 & & \\
Claude Opus~4.7 & 0.920 & & \\
\bottomrule
\end{tabular}}
\end{table}

\paragraph{A candidate mechanism.} We offer one explanation,
grounded in pre-training signal density: models with denser
exposure to entity names may recognize them more automatically
and confidently, making a visible system label harder to
override. This matches Claude Opus~4.7's pattern: highest
name-dependence stability on DY (JSR 0.812--0.937 across P1--P3 and P6
under Prompt~I, Table~\ref{tab:perturbation}) yet the largest BIO collapse under
anonymization (PBI 3.884, MRI 68.8\%), and in the human
evaluation (\S\ref{sec:human-eval}), the best label-free
agreement with human consensus ($\rho = 0.962$, MAE $= 0.12$) alongside
the largest label sensitivity under label-exposing prompts.
We treat this as an empirical regularity across the three
tested models, not a settled explanation; testing it on
open-source models is future work (\S\ref{sec:limitations}).

\paragraph{Generalization beyond entity alignment.} The
structural property producing anchor bias is not specific to
EA: any structured prediction task where a judge sees a
discrete system decision alongside its evidence creates the
same rationalization target. We expect this in relation
extraction, semantic role labeling, coreference resolution, and
slot filling, since each exposes a discrete, visible decision
to the judge in the same way EA does. This is a testable
prediction, not an established finding; validating it on at
least one additional task is the highest-priority direction for
follow-up work, and our released auditing framework is designed
to support exactly this extension (\S\ref{sec:limitations}).

\paragraph{Prompt~III scope (LLM as predictor).} Label-free evaluation succeeds on
distinctive-name datasets (DW) and with structure-aware EA systems
(NeuSymEA on both DW and DY), but fails on near-identical name
datasets paired with weak EA systems (DY EasyEA,
J-ROC-AUC 0.468--0.527, Score Gap $+$1.69--$+$2.25,
non-significant). This failure reflects the absence of lexical
discriminative signal when entity names are near-identical---a
limitation no prompt design can overcome without structural evidence.

\end{document}